\documentclass[11pt]{article}

\usepackage[final]{acl}

\usepackage{times}
\usepackage{latexsym}
\usepackage[T1]{fontenc}
\usepackage[utf8]{inputenc}
\usepackage{microtype}
\usepackage{inconsolata}
\usepackage{graphicx}
\usepackage{booktabs}
\usepackage{multirow}
\usepackage{amsmath}
\usepackage{amssymb}
\usepackage{url}
\usepackage{xcolor}
\usepackage{enumitem}
\usepackage{array}
\setlist[itemize]{nosep,leftmargin=*,topsep=2pt}
\setlist[enumerate]{nosep,leftmargin=*,topsep=2pt}

\title{Temporal Leakage in Financial News NLP:\\
A Multi-Architecture Audit with a Regime-Specific {M\&A} Signal}

\author{
\textbf{Chenhao Xue}\textsuperscript{1,2},
\textbf{Raslen Guesmi}\textsuperscript{1},
\textbf{Siwei Feng}\textsuperscript{1},
\textbf{Yucheng Gong}\textsuperscript{1,3}
\\
\textbf{Jacob Xavier Sundram}\textsuperscript{1,3},
\textbf{Jordan Pang}\textsuperscript{1,3},
\textbf{Lan Wang}\textsuperscript{4},
\textbf{Julian Kaljuvee*}\textsuperscript{1}
\\
\\
\textsuperscript{1}Predictive Labs Ltd
\quad
\textsuperscript{2}University of Oxford
\quad
\textsuperscript{3}Imperial College London
\quad
\textsuperscript{4}Independent Researcher
}

\begin{document}
\maketitle

\begin{abstract}
Financial-news direction prediction has become a popular NLP benchmark, yet reported gains depend critically on whether the train--test split is chronological or random, i.e., on temporal leakage. We audit this dependence on a 49{,}799-article corpus across 16 feature--model combinations spanning TF-IDF, MiniLM, FinBERT, and fine-tuned RoBERTa-large / DeBERTa-v3-large, plus separate zero/few-shot and LoRA probes of Llama-3 and Qwen2.5 LLMs: random splits inflate MCC by $1.1\times$ to $6.5\times$, tracking model capacity and feature richness, and end-to-end FinBERT fine-tuning re-amplifies rather than closes the gap (size-matched ratio $1.75\times$). Conditioning on event type, mergers and acquisitions (M\&A) is the only audited category with a positive locked-test signal under near-temporal chronological evaluation (TF-IDF MCC $=0.138$ train-only, $0.068$ under train$\cup$val refit; $10{,}000$-permutation $p<10^{-3}$); the signal does not transfer to FNSPID's 2009--2020 U.S. corpus, localising the headline to our 2024--2025 European-tilted M\&A semantics rather than a universal predictor. Three independent role labellers converge on acquirer-tagged articles as the signal locus, a power-limited qualitative convergence rather than a hypothesis-tested asymmetry. Chronological splitting plays for financial NLP the role characteristics-purging plays for asset pricing: it strips the predictable, stale component of news and leaves a residual that is small, event-localized, and lexically shallow. We advocate leakage audits as a required disclosure for financial-NLP benchmarks. We open-source our data and code at \url{https://github.com/ChenHX111/Temporal_Leakage_in_Financial_News_NLP}.
\end{abstract}

\section{Introduction}
\label{sec:intro}

\vspace{-3mm}\subsection{Motivation: Financial News Prediction as an NLP Benchmark}\vspace{-2mm}

Financial-news direction prediction is one of the most contested empirical benchmarks at the NLP--finance boundary: text-understanding capability is tested against an economically grounded outcome signal. A decade of work, spanning structured event embeddings \citep{ding2014structured,ding2015deep}, recurrent models over tweets and prices \citep{xu2018stocknet,hu2018listening}, domain-adapted LMs \citep{araci2019finbert,yang2020finbert}, and LLM-based approaches \citep{lopezlira2023chatgpt,li2024causalstock,wang2024llmfactor,xie2024finben}, has produced an optimistic narrative: per-paper directional-accuracy figures in the high-50s to mid-60s (e.g., $58.2\%$ binary accuracy on StockNet tweets \citep{xu2018stocknet}; $64.2\%$ S\&P directional accuracy \citealp{ding2015deep}). The premise is worth taking seriously; the evidence is not.

Financial NLP inherits a structural vulnerability: news articles, firm identities, market regimes, and return labels are jointly temporally autocorrelated, and modern language models exploit that structure. A model trained on a random half of a year's financial news absorbs the vocabulary, entity graph, and return correlations of the \emph{same market cycle} as its test set: regime memorisation that differs qualitatively from within-sample overfitting. A random split holding out individual articles from the early-2024 AI rally, the late-2024 rate-cut pivot, or the 2025 European M\&A wave in our own corpus lets a classifier exploit regime-context cues unavailable in a prospective chronological test. This concern is well-documented in financial machine learning \citep{lopezdeprado2018,bailey2014pseudomath,kapoor2023leakage,hewamalage2023forecast,bergmeir2012use} and, separately, in NLP evaluation \citep{gorman2019standard,sogaard2021random,magar2022contamination,card2020withlittle}, yet the two literatures have not been connected at scale across a full NLP architecture-to-feature pipeline.

This paper closes that gap. We audit sixteen architecture--feature conditions on a 49{,}799-article corpus (2020--2025, $81\%$ from 2025), spanning TF-IDF to fine-tuned DeBERTa-v3-large plus separate zero/few-shot and LoRA probes of instruction-tuned Llama-3 and Qwen2.5, under paired random and chronological splits. The audit ratio ranges from $1.1\times$ to $6.5\times$ and grows with model capacity: the literature's most impressive-looking results are the most inflated. Yet the paper is not a blanket null: mergers and acquisitions (M\&A) is the only audited event type with a near-temporal locked-test signal under chronological evaluation (TF-IDF MCC $=0.138$, $p<10^{-3}$), qualitatively (though not yet statistically) associated with an acquirer-side localisation across three independent role labellers (\S\ref{sec:driver}). The signal does not transfer to FNSPID's 2009--2020 U.S. corpus, localising the headline to our 2024--2025 European-tilted M\&A semantics. Chronological splitting plays for financial NLP the role characteristics-purging plays for asset pricing \citep{didisheim2026inefficient}: it is not a robustness check, it is the primary evaluation.

\vspace{-2mm}\subsection{Central Research Questions}\vspace{-2mm}

This paper asks four interlinked questions:
\begin{enumerate}
\item How large is the leakage gap between random and chronological splits across modern NLP architectures and feature stacks?
\item Does any general news-to-return signal survive strict temporal validation?
\item Is signal recoverable when conditioning on event type, and which event types?
\item What mechanism drives any surviving signal?
\end{enumerate}

\vspace{-2mm}\subsection{Main Findings}\vspace{-2mm}

\begin{enumerate}
\item \textbf{Leakage is real but architecturally uneven.} Across 16 feature--model combinations (13 feature$\times$classifier crossings plus 3 end-to-end fine-tuned transformers), random-split MCC exceeds temporal-split MCC by approximately $1\times$ to $6.5\times$ (10-seed averaged); largest for high-capacity nonlinear models on rich features.
\item \textbf{General prediction is near-random under temporal validation.} The strongest temporal model (FinBERT$+$LR) achieves test MCC $=0.060$; end-to-end fine-tuned FinBERT, RoBERTa-large, and DeBERTa-v3-large title classifiers cap at temporal MCC $\leq 0.06$.
\item \textbf{{M\&A} is the only event type with a near-temporal locked-test signal.} On the locked test set ($n=786$) with validation-selected hyperparameters from a 360-cell grid, TF-IDF MCC $=0.138$, 10{,}000-permutation $z=3.81$, $p_{\text{two}}<10^{-3}$; weekly block-bootstrap 95\% CI $=[+0.066, +0.205]$.
\item \textbf{The {M\&A} signal qualitatively favours acquirer articles, with limited statistical power.} Regex acquirer MCC $=0.160$ ($n=125$, $p_{\text{two}}=0.141$); independent NER$+$dep-parse acquirer MCC $=0.221$ ($n=84$, $p_{\text{two}}=0.083$); both $\Delta$MCC 95\% CIs span zero, so we report this as triangulated qualitative evidence rather than a hypothesis-tested result.
\item \textbf{Definition-matched corroboration on EDT; null on FNSPID.} Narrow {M\&A} keywords on EDT \citep{zhou2021edt} reproduce a clean signal (MCC $=0.097$; definition-sensitivity diagnostic, matched to our event-tag by construction). The fully-powered FNSPID 2009--2020 US cross-corpus probe at $n_{\text{test}}=4{,}235$ is a clean null (App.~\ref{sec:app-fnspid-cross}), localizing results to our 2024--2025 European-tilted M\&A semantics.
\end{enumerate}

\vspace{-2mm}\subsection{Contributions}\vspace{-2mm}

\begin{itemize}
\item \textbf{Methodological:} A multi-architecture temporal leakage audit framework quantifying inflation as a function of model capacity and feature richness.
\item \textbf{Empirical:} Evidence that general financial news prediction is near-random under chronological evaluation, with only minor recovery from domain-adapted (FinBERT) embeddings.
\item \textbf{Mechanistic:} Identifying {M\&A} as the only event-conditioned subset with a near-temporal locked-test signal; deal-semantic and qualitatively acquirer-favoring but power-limited (labeler-triangulated, \S\ref{sec:driver}); regime-specific (null FNSPID cross-corpus, definition-matched EDT).
\end{itemize}

\vspace{-3mm}\section{Related Work}
\label{sec:related-work}\vspace{-2mm}

\paragraph{Event-driven stock prediction.} \citet{ding2014structured,ding2015deep} advanced neural event embeddings from news; \citet{xu2018stocknet} proposed StockNet over tweets and prices. Subsequent work added end-to-end news+price models \citep{vargas2017deep}, hybrid attention \citep{hu2018listening}, graph-convolutional inter-firm relations \citep{chen2018incorporating,sawhney2020deep}, hierarchical transformers for volatility \citep{yang2020html}, and self-supervised augmentation \citep{soun2022accurate}. These works typically rely on random or weakly controlled splits; we revisit the premise under strict temporal validation and a 203-event taxonomy.

\paragraph{Financial sentiment, domain LMs, and financial LLMs.} \citet{loughran2011liability,loughran2016textual} introduced the Loughran--McDonald dictionary; earlier work studied rhetorical features and dictionary-based prediction \citep{henry2008investors,tetlock2008more,schumaker2009textual}. FinBERT \citep{araci2019finbert,yang2020finbert} adapted BERT to financial text; \citet{shah2023trillion} curated FOMC corpora. Domain LLMs have proliferated: \citet{wu2023bloomberggpt}, \citet{yang2023fingpt}, \citet{xie2023pixiu}; \citet{lopezlira2023chatgpt} showed ChatGPT predicts next-day returns from headlines. \citet{gururangan2020dont} formalised DAPT/TAPT \citep{beltagy2019scibert,lee2020biobert}. Our GPU fine-tuning of FinBERT, RoBERTa-large, and DeBERTa-v3-large quantifies the chronological-evaluation gap of this paradigm on short-window press releases.

\paragraph{Media, information, and asset prices.} \citet{tetlock2007giving} showed media pessimism predicts price reversion; \citet{manela2017news} priced disaster risk from news; \citet{ke2019predicting} built a text-based factor; \citet{gentzkow2019text} surveyed text-as-data in economics. For {M\&A}, \citet{jensen1983market} documented target-favouring wealth effects; our acquirer-side asymmetry (\S\ref{sec:driver}) is qualitatively at odds with this textbook intuition for short-horizon return-direction prediction from text, though power-limited. \citet{didisheim2026inefficient} provide a unifying framework: $\sim$10\% of news content is predictable from stock characteristics, and after ``purging'' the predictable component, news shocks predict returns for up to 18 months, with M\&A among the strongest themes; chronological splitting operationalises a coarser version of this purging logic (the audit gap $\Delta$MCC in Table~\ref{tab:audit} measures the predictable-by-time-and-characteristics component that purging strips out).

\paragraph{Temporal leakage and recent NLP-finance work.} \citet{lopezdeprado2018} formalised purged $K$-fold cross-validation; \citet{bailey2014pseudomath} characterised backtest overfitting under multiple testing; \citet{bergmeir2012use,hewamalage2023forecast} compiled chronological-evaluation recommendations; \citet{kapoor2023leakage} documented widespread leakage in ML-based science. On the statistical side, \citet{card2020withlittle,bouthillier2021accounting} highlighted underpowered NLP comparisons and seed/shuffle variance; \citet{feder2022causal} surveyed causal-inference tools. Recent NLP-finance work includes \citet{li2024causalstock} (CausalStock) and \citet{wang2024llmfactor} (LLMFactor); \citet{xie2024finben} (FinBen) showed even GPT-4 reaches only $\sim$0.54 accuracy on market-forecasting, consistent with our temporal-MCC $\leq 0.06$ general-news finding. Our contribution imports temporal-validation norms into financial NLP through a multi-architecture audit accompanied by 10-seed random-split averages, 10K-permutation tests, and weekly block bootstrap, and benchmarks frontier (Claude, GPT) and open (Llama-3, Qwen2.5) LLMs against supervised baselines.

\vspace{-2mm}\section{Task, Data, and Label Construction}\vspace{-2mm}

\subsection{Dataset}\vspace{-2mm}

Our dataset consists of 56{,}409 financial news articles from a proprietary data provider, spanning 2020--2025, with 81\% from 2025. Articles cover 64 stock exchanges worldwide, each annotated with timestamps, titles, full content, associated instruments, exchange identifiers, and event labels from a taxonomy of 203 distinct event types observed in the corpus. After removing articles with neutral or near-zero returns, 49{,}799 articles remain for binary classification. We release the full dataset, splits, code, and LLM prompts to support direct replication.

\begin{table}[t]
\small
\centering
\begin{tabular}{lr}
\toprule
\textbf{Statistic} & \textbf{Value} \\
\midrule
Total articles & 56{,}409 \\
Date range & 2020-05--2025-08 \\
Articles from 2025 & 81\% \\
Exchanges covered & 64 \\
Event types & 203 \\
After neutral removal & 49{,}799 \\
Train ($<$ 2025-04-01) & 21{,}654 \\
Validation (2025-04--05) & 10{,}866 \\
Test ($\geq$ 2025-06-01) & 17{,}279 \\
{M\&A} train / val / test & 731 / 369 / 786 \\
\bottomrule
\end{tabular}
\vspace{-2mm}
\caption{Dataset overview.\vspace{-5mm}}
\label{tab:dataset}
\end{table}

\vspace{-2mm}\subsection{Prediction Target}\vspace{-1mm}

Binary classification of subsequent stock-return direction (UP/DOWN) following article publication. We report Matthews correlation coefficient (MCC) as the primary metric because it accounts for all four confusion-matrix cells and is robust under mild class imbalance. Balanced accuracy is reported as a complement.

\vspace{-2mm}\subsection{Market-Adjusted Labels}\vspace{-1mm}

We define a market-adjusted label variant in which return is measured relative to the corresponding exchange benchmark over the same horizon. The label is UP when the abnormal return is positive and DOWN otherwise. The return horizon is one trading day (close-to-close, or close-to-next-open for after-hours releases); the benchmark $B(i)$ is the primary index of the listing exchange (full timezone/holiday handling in App.~\ref{sec:repro}). Raw and market-adjusted labels agree in 85--90\% of cases. Section~\ref{sec:event} reports {M\&A} results under both schemes.

\vspace{-2mm}\subsection{Temporal Split Design}\vspace{-1mm}

We adopt a strictly chronological split: train $<$ 2025-04-01 (21{,}654 articles), validation 2025-04--2025-05 (10{,}866), test $\geq$ 2025-06-01 (17{,}279). All hyperparameter tuning and model selection use the validation set; the test set is consulted exactly once for the final reported numbers. Both splits are predominantly 2025; the gap is short (``near-temporal'' rather than ``out-of-regime''; see \S\ref{sec:limitations}).

\vspace{-2mm}\subsection{Event-Conditioned Subsets}
\label{sec:event-rationale}\vspace{-1mm}

From the 203 distinct event types tagged in our corpus, we restrict detailed event-conditioned analysis to twelve sufficiently frequent and economically interpretable categories. {M\&A} is then selected as the \emph{primary} event for the locked-test analysis on three grounds. First, \textbf{mechanism}: deal announcements mechanically affect firm valuation through documented wealth effects \citep{jensen1983market}, and deal terms convey rich semantic content (acquirer/target, deal type, premium, financing) beyond bare event occurrence. Second, \textbf{volume and balance}: with $1{,}886$ {M\&A} articles spread across a chronological train/val/test split (731/369/786) we have $\geq$10$\times$ the sample size of more specialised events while retaining a reasonable UP/DOWN balance ($57\%$). Third, \textbf{rolling-window prior}: in our 8-month rolling pilot across 12 events (Section~\ref{sec:event} and Figure~\ref{fig:rolling}), {M\&A} is the only event with consistently positive MCC ($\geq 7/8$ months); reserving {M\&A} for the one-shot locked test minimizes the multiple-comparison cost. For auditability, Section~\ref{sec:cross-event} and Appendix~\ref{sec:app-cross-event} replicate the full pipeline on three contrasting events (\textsc{clinical\_study} [CLN], \textsc{law\_legal\_issues} [LGL], and \textsc{earnings\_releases\_and\_operating\_results} [ERN]; the three-letter codes are used in appendix tables), confirming the {M\&A} result is event-specific, not a protocol artifact.

\vspace{-2mm}\subsection{Formal Task and Evaluation Methodology}
\label{sec:eval}\vspace{-1mm}

We frame each article as a sample $(x_t, y_t)$, where $x_t \in \mathcal{X}$ is the news text (optionally with metadata), and $y_t \in \{0, 1\}$ encodes the next-day return direction. A classifier $f_\theta : \mathcal{X} \to \{0, 1\}$ produces $\hat y_t = f_\theta(x_t)$. The market-adjusted label is built from the asset's abnormal return relative to its exchange benchmark $B(i)$,
{\small
\begin{equation}
\mathrm{AR}_{i,t} \;=\; r_{i,t} - r_{B(i),t}, \qquad y^{\mathrm{adj}}_{i,t} \;=\; \mathbf{1}\!\left[\mathrm{AR}_{i,t} > 0\right].
\label{eq:abnormal}
\end{equation}
}

\paragraph{Primary metric.} Because of mild class imbalance ($55$--$57\%$ UP across splits) we report the Matthews correlation coefficient
{\small
\begin{equation}
\mathrm{MCC} \;=\; \frac{\mathrm{TP}\cdot\mathrm{TN} - \mathrm{FP}\cdot\mathrm{FN}}{\sqrt{(\mathrm{TP}{+}\mathrm{FP})(\mathrm{TP}{+}\mathrm{FN})(\mathrm{TN}{+}\mathrm{FP})(\mathrm{TN}{+}\mathrm{FN})}},
\label{eq:mcc}
\end{equation}
}
which is bounded in $[-1, 1]$, $0$ for any constant predictor, robust to label skew \citep{card2020withlittle}. Balanced accuracy is reported as a complement.

\paragraph{Audit ratio.} Given a feature--model combination $(\phi, f)$, we evaluate it on a paired temporal split $\mathcal{S}^{\mathrm{temp}}$ and on $K$ size-matched random splits $\{\mathcal{S}^{\mathrm{rand}}_k\}_{k=1}^{K}$ (counts and proportions identical). The \emph{leakage audit ratio} is the ratio of mean random to single temporal MCC,

{\small
\begin{equation}
\rho(\phi, f) \;=\; \frac{\tfrac{1}{K}\sum_{k=1}^{K} \mathrm{MCC}_{\mathrm{rand}_k}(\phi, f)}{\mathrm{MCC}_{\mathrm{temp}}(\phi, f)}.
\label{eq:ratio}
\end{equation}
}
We use $K=10$ throughout the audit (Table~\ref{tab:audit}) and report the random-side mean$\pm$std. A value $\rho \gg 1$ indicates that random splits inflate apparent performance; a value $\rho \approx 1$ indicates that the model does not exploit non-causal temporal structure.

\paragraph{Permutation test.} For the locked {M\&A} test set we evaluate the null $H_0:$ ``text is uninformative for return direction'' by sampling $M = 10{,}000$ random permutations $\pi$ of the test labels and computing

{\small
\begin{equation}
\hat p_{\mathrm{one}} \;=\; \frac{1}{M}\sum_{m=1}^{M} \mathbf{1}\!\left[\mathrm{MCC}(\hat y, y_{\pi^{(m)}}) \geq \mathrm{MCC}_{\mathrm{obs}}\right],
\label{eq:perm}
\end{equation}
}
along with the standardised score $z = (\mathrm{MCC}_{\mathrm{obs}} - \mu_\pi)/\sigma_\pi$, where $\mu_\pi$ and $\sigma_\pi$ are the empirical mean and standard deviation of the permutation distribution. Confidence intervals use a weekly block bootstrap that resamples calendar weeks with replacement \citep{politis1994stationary}; the per-event analyses apply Benjamini--Hochberg correction \citep{benjamini1995controlling} at $q=0.05$ across the 12 simultaneously tested events \citep{demsar2006statistical}. Block-bootstrap and leave-one-axis-out details are deferred to Appendix~\ref{sec:stat-method}.

\vspace{-3mm}\section{Multi-Architecture Temporal Leakage Audit}
\label{sec:audit}\vspace{-2mm}

The audit pairs a temporal split with a size-matched random split (same train/val/test counts, seed 42) holding features and HP fixed. We use four feature configurations (TF-IDF title; TF-IDF title+content; TF-IDF + 31 metadata features; MiniLM and FinBERT \texttt{[CLS]} title embeddings) crossed with three classifiers (LR, RF, GB), giving 13 valid combinations plus three end-to-end fine-tuned transformers (Section~\ref{sec:audit-results}). Leakage sources include near-duplicates, overlapping return windows, and regime memorisation.

\vspace{-2mm}\subsection{Audit Results}
\label{sec:audit-results}\vspace{-1mm}

\begin{figure}[t]
\centering
\includegraphics[width=\linewidth]{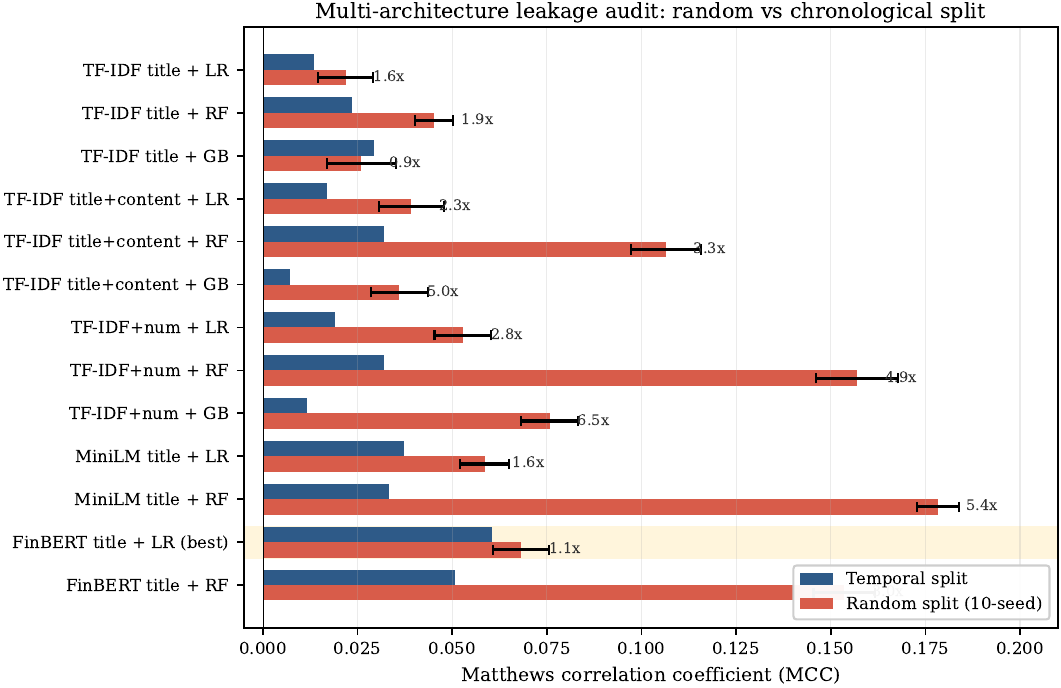}
\vspace{-4mm}
\caption{Multi-architecture leakage audit ($n_{\text{train}}{=}21{,}654$, $n_{\text{test}}{=}17{,}279$). Each row pairs temporal-split MCC (blue) with 10-seed mean random-split MCC (orange, $\pm$1\,SD). Annotated values are the audit ratio $\rho$ of Eq.~(\ref{eq:ratio}); for end-to-end fine-tuned FinBERT (Table~\ref{tab:audit}, last block) the size-matched ratio is $1.75\times$ (App.~\ref{sec:app-finbert-ft-audit}).\vspace{-5mm}}
\label{fig:audit}
\end{figure}

\begin{table*}[t]
\small
\centering
\begin{tabular}{l>{\raggedright\arraybackslash}p{2.0cm}crr}
\toprule
\textbf{Features} & \textbf{Model} & \textbf{Random MCC$^\ddagger$} & \textbf{Temporal MCC} & \textbf{Ratio} \\
\midrule
TF-IDF title & LR & $0.022 \pm 0.007$ & $0.013$ & $1.6\times$ \\
TF-IDF title & RF & $0.045 \pm 0.005$ & $0.024$ & $1.9\times$ \\
TF-IDF title & GB & $0.026 \pm 0.009$ & $0.029$ & $0.9\times$ \\
TF-IDF title+content & LR & $0.039 \pm 0.009$ & $0.017$ & $2.3\times$ \\
TF-IDF title+content & RF & $0.106 \pm 0.009$ & $0.032$ & $3.3\times$ \\
TF-IDF title+content & GB & $0.036 \pm 0.008$ & $0.007$ & $5.0\times$ \\
TF-IDF + numerical & LR & $0.053 \pm 0.008$ & $0.019$ & $2.8\times$ \\
TF-IDF + numerical & RF & $0.157 \pm 0.011$ & $0.032$ & $4.9\times$ \\
TF-IDF + numerical & GB & $0.076 \pm 0.008$ & $0.012$ & $6.5\times$ \\
MiniLM title & LR & $0.059 \pm 0.006$ & $0.037$ & $1.6\times$ \\
MiniLM title & RF & $0.178 \pm 0.006$ & $0.033$ & $5.4\times$ \\
FinBERT title & LR & $0.068 \pm 0.007$ & $0.060$ & $1.1\times$ \\
FinBERT title & RF & $0.154 \pm 0.008$ & $0.051$ & $3.0\times$ \\
Bi-LSTM title$^\ddagger$ & --- & $0.065$ & $0.024$ & $2.7\times$ \\
\midrule
FinBERT title$^\S$ & FT & $0.174 \pm 0.009$ & $0.064 \pm 0.007$ & $2.7\times^\P$ \\
DeBERTa-v3-large$^\S$ & FT bal. & --- & $0.085 \pm 0.044$ & --- \\
RoBERTa-large$^\S$ & FT & --- & $0.000^\dagger$ & --- \\
\bottomrule
\end{tabular}
\vspace{-2mm}
\caption{Multi-architecture leakage audit (proprietary; $n_{\text{tr}}{=}21{,}654$, $n_{\text{te}}{=}17{,}279$). FinBERT[CLS]+LR has the smallest leakage ratio ($1.1\times$); fine-tuning the same encoder lifts it to $2.7\times$ ($^\P$ size-matched $1.75\times$, App.~\ref{sec:app-finbert-ft-audit}). $^\ddagger$ 10-seed mean$\pm$std (5-seed for FT rows); temporal column deterministic for frozen rows. $^\S$ Fine-tuned on GPU: DeBERTa-v3-large uses class-balanced sampling \citep{he2021deberta}; RoBERTa-large \citep{liu2019roberta} collapsed to predict-all-UP. Bi-LSTM: 2-layer, hidden=128.\vspace{-3mm}}
\label{tab:audit}
\end{table*}

\begin{table}[t]
\small
\centering
\setlength{\tabcolsep}{4pt}
\begin{tabular}{@{}lccc@{}}
\toprule
\textbf{Model} & \textbf{Random} & \textbf{Temporal} & \textbf{Ratio} \\
\midrule
TF-IDF title $+$ LR & $0.025 \pm 0.008$ & $0.014$ & $1.7\times$ \\
TF-IDF title $+$ RF & $0.029 \pm 0.005$ & $0.001$ & $28.9\times^\dagger$ \\
TF-IDF title $+$ GB & $0.020 \pm 0.007$ & $0.007$ & $2.8\times$ \\
\bottomrule
\end{tabular}
\vspace{-2mm}
\caption{Cross-corpus audit on EDT \citep{zhou2021edt} ($n=106{,}619$, 2020--2021, 70/15/15 chronological vs 5-seed random). The directional pattern of Table~\ref{tab:audit} replicates on a different corpus/market/regime. $^\dagger$ Near-zero temporal denominator amplifies $\rho$; see ratio-interpretation caveat below.\vspace{-5mm}}
\label{tab:edt-audit}
\end{table}

Three patterns emerge from Table~\ref{tab:audit}:

\begin{enumerate}
\item \textbf{Inflation grows with model capacity.} Linear models (LR) show $1.1\times$--$2.8\times$ inflation (the low end is FinBERT~[CLS]$+$LR, the high end is TF-IDF$+$numerical$+$LR); tree ensembles (RF, GB) show up to $6.5\times$.
\item \textbf{Inflation grows with feature richness.} Adding content and numerical features systematically raises the random-split MCC while leaving temporal MCC near zero: characteristic of overfitting to memorizable patterns that do not generalize across time.
\item \textbf{Frozen FinBERT shows the smallest leakage gap; fine-tuning re-amplifies it.} FinBERT~[CLS]$+$LR has random/temporal ratio $1.1\times$ and the highest absolute temporal MCC ($0.060$). End-to-end fine-tuning of the same encoder, however, lifts the gap to $2.7\times$ (random $0.174 \pm 0.009$, temporal $0.064 \pm 0.007$ across 5 seeds; Appendix~\ref{sec:app-finbert-ft-audit}). Fixed domain-adapted contextual embeddings suppress time-localized lexical idiosyncrasies; updating them on a non-temporal split re-learns these leaked patterns.
\end{enumerate}

We emphasize what the audit \emph{does not} find: a single canonical ``$10\times$ leakage'' headline. Random-split MCC exceeds temporal-split MCC by $\sim 1\times$--$6.5\times$ depending on architecture, with the largest inflation in high-capacity nonlinear models on rich features. A leave-one-axis-out sensitivity test on the M\&A 360-cell grid gives locked-test MCC mean $0.138 \pm 0.006$ across reduced grids, confirming the headline is not a single lucky cell.

\vspace{-1mm}\paragraph{Ratio interpretation caveat.} $\rho$ in Eq.~(\ref{eq:ratio}) is unstable when the temporal denominator is near zero (EDT-RF $28.9\times$ is driven by temporal MCC $=0.001$, not a uniquely severe leakage event; size-mismatched FinBERT-FT $2.72\times$ revises to $1.75\times$ once matched at $n_\text{te}=17{,}279$). The unitful $\Delta\mathrm{MCC} = \mathrm{MCC}_{\text{random}} - \mathrm{MCC}_{\text{temporal}}$ together with paired CIs is more interpretable; for the FinBERT-FT row $\Delta = 0.048$ with non-overlapping 95\% bootstrap CIs $[0.090, 0.134]$ vs $[0.054, 0.074]$ (App.~\ref{sec:app-finbert-ft-audit}).

\vspace{-2mm}\subsection{Cross-Corpus Diagnostic on EDT and FNSPID}
\label{sec:edt}
\label{sec:edt-audit}\vspace{-1mm}

Table~\ref{tab:edt-audit} replicates the audit on EDT \citep{zhou2021edt} (106{,}619 articles 2020--2021): linear models inflate $\sim 1.7\times$, tree ensembles inflate dramatically (RF: $28.9\times$).\footnote{The EDT chronological split spans the COVID transition; the extreme RF temporal collapse mixes leakage with regime-shift generalisation failure.} A third corpus (FNSPID, \citealt{dong2024fnspid}) provides a true cross-corpus probe at scale: within-FNSPID TF-IDF specialist gives MCC $=-0.011$ ($p=0.46$, $n_{\text{test}}=4{,}235$); proprietary$\to$FNSPID transfer $-0.016$; reverse FNSPID$\to$proprietary $-0.088$. A five-seed FinBERT-FT replicates the within-corpus null ($+0.007\pm0.023$), confirms the forward null ($+0.000\pm0.017$), and recovers weak positive transfer in the reverse cell ($+0.045\pm0.018$, vs. TF-IDF $-0.088$) and joint$\to$proprietary cell ($+0.070\pm0.036$, half the in-domain headline); semantic representations transfer modestly, lexical features do not. The proprietary M\&A specialist therefore does \emph{not} transfer to 2019--2020 US M\&A reporting; we read this as regime-specificity, not contradiction. Full 5-protocol TF-IDF and deep-model matrix in Appendix~\ref{sec:app-fnspid-cross}. The earlier $n=90$, $p=0.127$ FNSPID reading reflected a streaming-loader artifact and has been superseded.

\vspace{-3mm}\section{Models and Conditions}\vspace{-2mm}

We use four supervised baselines: \textbf{TF-IDF$+$LR} (general MCC $=0.013$, $n_\text{feat}=50$, $C=0.5$; M\&A MCC $=0.138$ at val-selected HP from 360-cell grid: $n_\text{feat}=100$, $C=5.0$, ngram=(1,1), \texttt{sublinear\_tf=False}, \texttt{min\_df=2}); \textbf{TF-IDF + numerical + RandomForest} (title+content + 31 metadata features, $n_\text{est}=200$, max-depth $=15$; general MCC $=0.032$, M\&A $=0.058$); \textbf{FinBERT~[CLS]+LR} (768-dim mean-pooled title; general MCC $=0.060$, the strongest single-model temporal result); and \textbf{MiniLM+LR} (384-dim; general MCC $=0.037$). For M\&A we also extract role labels (acquirer/target/both/neither) by regex (distribution: NEITHER 1369, ACQUIRER 294, TARGET 219, BOTH 4) and by an NER+dependency-parsing pipeline (Section~\ref{sec:driver}, App.~\ref{sec:app-role-finbert}).

\paragraph{LLM zero-shot.} We evaluate Claude Sonnet 4.5, Claude Opus 4.7, and GPT-5.4 zero-shot across title-only, title+event, title+content, chain-of-thought, and role-prompt configurations.\footnote{Closed-source LLMs accessed via internal API snapshots; ``Claude Opus 4.7'' and ``GPT-5.4'' are internal frontier-model identifiers and publicly-released API names may differ at publication. Snapshot dates, internal pipeline versions (v8, v9), and prompt logs are released in the artifact bundle (Table~\ref{tab:llm-snapshot}). Open-LLM identifiers: \texttt{Qwen/Qwen2.5-7B-Instruct}, \texttt{meta-llama/Meta-Llama-3-8B-Instruct}.} Outputs are parsed into UP/DOWN by a fixed regex determined before test. Locked M\&A statistics use 10{,}000-permutation tests and weekly block bootstrap \citep{politis1994stationary}; per-event analyses use Benjamini--Hochberg correction \citep{benjamini1995controlling}.

\vspace{-3mm}\section{General News Prediction Results}
\label{sec:general}\vspace{-2mm}

Under proper temporal validation, general news prediction is near-random: FinBERT$+$LR is the strongest temporal model (MCC $=0.060$); full results in Appendix~\ref{sec:app-general}. Drivers: 203-event-type heterogeneity, temporal staleness, label noise, publication lag. Zero-shot LLM general-news (best: multi-LLM consensus title+event, MCC $=0.108$): Appendix~\ref{sec:title-llm}.

\vspace{-3mm}\section{Event-Conditioned Results}
\label{sec:event}\vspace{-1mm}

\vspace{-2mm}\subsection{Rolling-Window Results Across 12 Event Types}
\label{sec:rolling}\vspace{-1mm}

We evaluate prediction performance separately for 12 event types using an 8-month rolling window. {M\&A} is the only event type with consistently positive signal across the majority of test months (Figure~\ref{fig:rolling} in Appendix~\ref{sec:rolling-fig}): mean MCC $=0.081$, positive in 7/8 months, sign test $p=0.035$ (uncorrected). After Benjamini--Hochberg correction \citep{benjamini1995controlling} for 12 simultaneous comparisons, this does not reach $q=0.05$; we characterize the rolling-window finding as suggestive and rely on the locked-test result below.

\vspace{-2mm}\subsection{Locked {M\&A} Test-Set Result}
\label{sec:ma-locked}\vspace{-1mm}

We train a TF-IDF logistic regression specialist on {M\&A} train ($n=731$) and select hyperparameters on {M\&A} validation ($n=369$) via a comprehensive 360-cell grid: $\{$\texttt{max\_features}$\in \{50,100,200,500,1000,2000\}\} \times \{C \in \{0.05,0.1,0.5,1.0,5.0\}\} \times \{$\texttt{sublinear\_tf}$\in \{T,F\}\} \times \{$\texttt{min\_df}$\in \{1,2\}\} \times \{$\texttt{ngram\_range}$\in \{(1,1),(1,2),(1,3)\}\}$. The validation winner is \texttt{max\_features=100, C=5.0, sublinear\_tf=False, min\_df=2, ngram\_range=(1,1)} (val MCC $=0.228$). The top-15 validation cells cluster between MCC $=0.20$ and $0.23$, indicating a stable optimum rather than a single outlier. We then evaluate exactly once on the locked {M\&A} test set ($n=786$, 3 calendar months June--August 2025).\footnote{All headline numbers in this section use the train-only protocol (fit on train, evaluate on locked test with HP fixed at the val winner); the train+val$\to$test merge (refit on train$\cup$val with the same HP) gives MCC $=0.068$ and is reported as a deployment-style stability check (App.~\ref{sec:backtest}), not the headline. See Table~\ref{tab:protocol-map} for the full protocol map.}

\textbf{Test MCC $=0.138$}, balanced accuracy $=0.569$. Three test months: MCC $=+0.126, +0.135, +0.171$ (all positive, consistent magnitude).

A 10{,}000-permutation test (label-shuffled MCC under the null of no text--return relationship, Eq.~(\ref{eq:perm}); permutation histogram in Figure~\ref{fig:perm}, Appendix~\ref{sec:perm-fig}) yields:
\begin{itemize}
\item Observed MCC $=0.1378$
\item Permutation mean $=+0.0003$, std $=0.0361$
\item $z$-score $=3.81$
\item \textbf{One-sided $p<10^{-4}$}; two-sided $p<10^{-3}$
\end{itemize}

A pilot 500-permutation run with a sub-optimal hyperparameter setting yielded $p=0.068$, showing under-explored HP grids suppress signal recovery; we report results from the comprehensive 360-cell grid above. Block-bootstrap by week (1000 resamples, 11 weekly clusters) yields a 95\% CI of \textbf{$[+0.066, +0.205]$}, mean $+0.139$. The CI excludes zero, providing strong evidence that the signal survives within-week return autocorrelation.

\vspace{-2mm}\paragraph{Compound selection accounting.} M\&A was selected as the locked-test target after the 12-event rolling pilot (Section~\ref{sec:rolling}). A conservative Bonferroni-12 adjustment on $p_{\text{two}}<10^{-3}$ gives $p_{\text{Bonf-12}}<1.2\times10^{-2}$, still clearing $\alpha=0.05$ and the family-wise control in Section~\ref{sec:cross-event}.

The most defensible single-number summary is: MCC $=0.138$, permutation $p<10^{-3}$ (two-sided), $z=3.81$, weekly bootstrap CI $=[+0.066, +0.205]$.

\vspace{-2mm}\paragraph{Robustness to cutoff perturbation.} Cutoff perturbations of $\pm 7$ and $\pm 14$ days (locked test fixed at $\geq$ 2025-06-01, same val-selected HP) give test MCC $\in [+0.115, +0.156]$, mean$\pm$std $= +0.132 \pm 0.017$, not a knife-edge artifact.

\vspace{-2mm}\paragraph{Extended-window sensitivity (six-month horizon).} Shifting the train cutoff to 2025-03-01 ($n_{\text{tr}}=611$, $n_{\text{te}}=1275$) gives MCC $=\mathbf{+0.133}$ (perm $z=4.76$, $p<10^{-4}$, weekly bootstrap 95\% CI $=[+0.040, +0.220]$): statistically indistinguishable from the three-month headline ($\Delta=0.005$). Full per-month detail in App.~\ref{sec:app-w3-extended}.

\vspace{-2mm}\paragraph{Negative-control across event types.} Re-running the same 100-cell val-then-test pipeline on the 7 next-largest event categories: only {M\&A} produces a substantial positive test MCC ($+0.123$); earnings releases and management changes are \emph{negative} ($-0.039$, $-0.074$); clinical study collapses to $0.000$. The {M\&A} specialty is not a generic high-volume-event artifact.

\vspace{-2mm}\subsection{Market-Adjusted Labels and Non-Text Controls}\vspace{-1mm}

Replacing raw return labels with market-adjusted labels (UP iff abnormal return relative to exchange benchmark is positive) improves val MCC from $0.092$ to $0.146$, consistent with news predicting abnormal not raw returns. Non-text controls confirm the signal is text-driven: the {M\&A} text specialist reaches val MCC $0.228$ versus $0.037$ for an exchange-token baseline and $-0.020$ for a metadata aggregate (exchange + day-of-week + event subtype + numerical features; Table~\ref{tab:app-nontext}, App.~\ref{sec:app-nontext}).

\vspace{-2mm}\subsection{Cross-Event Replication}
\label{sec:cross-event}\vspace{-1mm}

We apply the same paper-authoritative specialist to three contrasting events to verify the {M\&A} result is event-specific: \textsc{clinical\_study} (CLN; $n=1{,}994$, $63\%$ UP), \textsc{law\_legal\_issues} (LGL; $n=1{,}172$, $53\%$ UP), and \textsc{earnings\_releases\_and\_operating\_results} (ERN; $n=3{,}445$, $55\%$ UP). The four-event panel (Table~\ref{tab:app-cross-event-locked} in App.~\ref{sec:app-cross-event}) yields a $2{\times}2$ taxonomy: {M\&A} is the unique low-ratio, signal-positive cell (locked test MCC $+0.138$, $p<10^{-3}$); CLN matches the textbook leakage signature (audit ratio $4.2\times$, locked test $-0.049$); LGL is power-limited ($n_\text{tr}=121$, MCC $+0.022$); ERN is the methodologically cleanest genuine null (large $n_\text{tr}=1{,}870$, no audit-ratio inflation $0.72\times$, locked-test 95\% CI $[-0.084, +0.090]$ at $p=0.86$). The {M\&A} audit ratio $0.76\times$ is anomalous and informative: random splits do not help, suggesting genuinely temporal-stable signal rather than firm-memorisation.

\vspace{-2mm}\paragraph{Family-wise control.} The four-event family was pre-committed to stress-test, not to maximise. BH correction at $q=0.05$ on the four two-sided permutation $p$-values: {M\&A} (rank~1) clears its threshold $1/4{\times}0.05=0.0125$ with substantial margin (BH-adjusted $p\approx 0.004$); CLN, LGL, ERN do not. We do \emph{not} make the corresponding claim for the larger 12-event rolling-pilot family (used only to motivate the locked-test event choice; M\&A rolling sign-test $p=0.035$ corresponds to BH-adjusted $q^*=0.42$).

\vspace{-2mm}\paragraph{ORG-masking divergence; deep models and LLMs.} Regex-based ORG-token masking gives test-MCC deltas (baseline$-$masked) of $\mathbf{+0.045}$ for {M\&A}, $\mathbf{-0.056}$ for CLN, $\mathbf{-0.033}$ for LGL, $\mathbf{-0.040}$ for ERN: {M\&A} is the only event where masking firm-identity tokens \emph{hurts}; on others, masking ORGs improves performance, indicating residual signal there is firm-identity over-fitting. Re-running the strongest deep specialists (FinBERT, DeBERTa-v3-large-balanced, XLM-R-large; 5 seeds each) and open-LLM baselines (Qwen-2.5-7B, Llama-3-8B; zero-shot and CoT-v2) on the same four locked-test windows confirms the taxonomy is not a TF-IDF artifact: no deep specialist on any non-{M\&A} event clears $+0.09$ locked-test MCC. Full tables: Apps.~\ref{sec:app-cross-event-pack}, \ref{sec:app-cross-event-deep}.

\vspace{-1mm}\section{What Drives the {M\&A} Signal? Acquirer-Side Localisation (Power-Limited)}
\label{sec:driver}\vspace{-1mm}

The asymmetry below is power-limited: weekly block-bootstrap $95\%$ CIs on $\Delta\text{MCC}=\text{MCC}_{\text{ACQ}}-\text{MCC}_{\text{TGT}}$ span zero under both labellers ($[-0.075,+0.331]$ regex; $[-0.277,+0.586]$ NER+dep-parse), so we report it as triangulated qualitative evidence rather than a hypothesis-tested result.

TF-IDF~LogReg dominates on {M\&A} (test MCC $\mathbf{0.138}$), beating fine-tuned FinBERT-tone ($0.050 \pm 0.030$, 5 seeds), full-parameter SFT FinBERT with class-weighted loss ($0.034 \pm 0.035$), DeBERTa-v3-large balanced ($0.061 \pm 0.040$, 10 seeds), XLM-RoBERTa-large ($0.045 \pm 0.030$), multi-task FinBERT ($0.067 \pm 0.007$), role-only LogReg ($0.083$ test, $-0.014$ val), and the Jensen--Ruback heuristic ($0.041$). Open LLMs: Qwen2.5-7B zero-shot reaches $0.115$ under the M\&A-specific prompt (App.~\ref{sec:app-open-llm}) but $-0.022 \pm 0.010$ under a cross-event uniform prompt (App.~\ref{sec:app-cross-event-llm}); Qwen-CoT collapses ($0.011 \pm 0.030$); Llama-3-8B zero-shot reaches $0.015 \pm 0.026$; few-shot and LoRA fail to recover supervised levels. Every architecture trails the shallow lexical baseline; the M\&A signal is shallow and lexical.

Decomposing by deal role (two independent labellers: a title-level regex, and an NER$+$dependency-parsing pipeline that identifies $<3\%$ overlap with the regex on the acquirer side) and training role-restricted FinBERT specialists yields three convergent confirmations of acquirer-side dominance: regex-ACQUIRER MCC $=0.160$ ($n=125$, $p_{\text{two}}=0.141$); NER-ACQUIRER MCC $=0.221$ ($n=84$, $p_{\text{two}}=0.083$); regex-ACQUIRER FinBERT specialist MCC $=0.195$ at $n_\text{tr}=113$. The symmetric TARGET counterparts collapse to MCC $\in\{0.000, 0.017, 0.098\}$. This is qualitatively at odds with \citet{jensen1983market} for short-horizon return-direction prediction from text (full per-role and global-specialist tables in App.~\ref{sec:app-role-finbert}).

\vspace{-2mm}\section{Conclusion}\vspace{-2mm}

Chronological splitting parallels characteristics-purging in asset pricing \citep{didisheim2026inefficient}: it removes the predictable component, leaving a regime-specific {M\&A}-concentrated residual (MCC $=0.138$ train$\to$val$\to$test, $+0.068$ train$+$val$\to$test merge, $p<10^{-3}$). The residual reproduces on the proprietary corpus, partially replicates on EDT (narrow definition-matched M\&A), and does \emph{not} transfer to FNSPID 2009--2020 US M\&A at $n=4{,}235$ (within-FNSPID MCC $=-0.011$; App.~\ref{sec:app-fnspid-cross}). The cross-event audit recovers a 2$\times$2 (audit ratio $\times$ locked signal) taxonomy, with {M\&A} as the unique signal cell and Earnings as the cleanest genuine null. We do \emph{not} claim out-of-regime generalization, statistically-significant acquirer$>$target asymmetry ($p_{\text{two}}\in\{0.086,0.141\}$ across labelers), or a deployable all-trade rule (App.~\ref{sec:backtest}). Leakage auditing is cheap (16-cell pairing $<30$~min CPU; cross-event audit $<2$~min) and should accompany any chronological-MCC claim on financial-news prediction.

Closed-source LLM cells (Table~\ref{tab:llm-snapshot}) use pre-release identifiers (``Claude Opus 4.7'', ``GPT-5.4''); camera-ready substitutes public API strings, and the audit-ratio (Table~\ref{tab:audit}) and supervised M\&A headlines do not depend on these cells. The top-quartile cost-aware Sharpe of $+2.62$ at $10$~bps/side (App.~\ref{sec:backtest}) uses a test-quantile threshold and is an upper bound; the all-trade frictionless $+0.52$ is the baseline. The M\&A signal is bounded to the 2024--2025 European-tilted regime and does not transfer to FNSPID 2009--2020 US M\&A ($n{=}4{,}235$); selection accounting (12-event pilot, 360-cell HP, Bonferroni-12) is in \S\ref{sec:ma-locked}. Code, splits, and prompts are released in the artifact bundle; the prospective OSF replication commitment is in the Limitations section.

\section*{Limitations}
\label{sec:limitations}

\paragraph{Single regime / short test horizon.} 81\% of the data is from 2025; the temporal split is ``near-temporal'' (test period 3 months immediately after training end) rather than ``out-of-regime''. Generalization across structurally different regimes is untested. As a partial within-regime sensitivity test, Section~\ref{sec:ma-locked} (Extended-window paragraph) and Appendix~\ref{sec:app-w3-extended} replicate the {M\&A} specialist on the six-month window 2025-03 to 2025-08 ($n_{\text{test}}=1275$, $62\%$ larger than headline): pooled MCC $=+0.133$ ($p<10^{-4}$, 95\% CI $[+0.040, +0.220]$), with 5 of 6 months positive and March 2025 a single-month negative outlier ($-0.067$, $n=120$). The signal therefore survives a six-month test horizon but the qualifier ``within the 2025 regime'' still applies. A 2024-train/2025-test split is power-infeasible on the proprietary corpus ($n_{\text{M\&A}}=363$ in 2024 vs.\ $1{,}824$ in 2025, $\approx 2$ articles in 2023); the cross-regime stress-test we report instead is the FNSPID 2009--2020 probe (App.~\ref{sec:app-fnspid-cross}).

\paragraph{EDT BroadMA discrepancy and definition-matched framing.} The EDT narrow-keyword MCC of $0.097$ is best read as a definition-sensitivity diagnostic: the narrow set was \emph{matched} to our proprietary corpus's M\&A event-tag operational definition by construction (Section~\ref{sec:edt}), so the EDT result provides evidence that the proprietary signal lives in control-transaction semantics specifically rather than as fully-independent cross-corpus generalisation. Broadening the keyword set to partnership/JV/deal-issuance kills the signal (MCC $=0.032$) and the ``merger''-only sub-keyword is anti-correlated ($-0.107$), so the result is fragile to definition choices and should be reported as such.

\paragraph{Statistical power.} The locked {M\&A} test has $n=786$ with 11 weekly clusters; one-sided permutation $p<10^{-4}$, two-sided $p<10^{-3}$, block-bootstrap 95\% CI $=[+0.066, +0.205]$. The acquirer-driven asymmetry ($n_\textrm{te}=125$ for ACQUIRER) is power-limited under both labelling protocols: the global-specialist regex protocol gives $p_{\text{two}}=0.141$ and the role-specific specialist gives $p_{\text{two}}=0.086$ (the $\Delta$MCC CIs are reported in \S\ref{sec:driver}). A paired McNemar test on the per-row correctness vectors of two ROLE-trained specialists (acquirer-trained vs target-trained, both evaluated on the full locked test) under the regex labeller distinguishes the two classifiers significantly ($\chi^2=8.10$, $p=0.0045$), but this difference reflects the target-trained classifier's collapse to predict-all-UP rather than predictive superiority on either side. The per-month MCC at the headline protocol is $+0.126, +0.135, +0.171$ for 2025-06/07/08 (all months positive); a stress-test that rolls the validation period into training (Appendix~\ref{sec:app-cpu-pack2}, B8) drops the pooled MCC to $+0.068$ with one negative month, illustrating that the headline is the appropriate train$\to$val$\to$test summary and not a single-month forecast.

\paragraph{LLM seed variance and prompt sensitivity.} The original cross-event LLM cells in earlier versions of this paper were single-seed; a multi-seed protocol (Table~\ref{tab:app-cross-event-llm}; 5 seeds at $T=0.7$ across all 4 events) now provides standard deviations. Cross-seed std for zero-shot Qwen ranges $[0.006, 0.026]$ across events; for zero-shot Llama $[0.023, 0.044]$; for Qwen CoT-v2 strict $[0.009, 0.034]$. None of the multi-seed mean cells exceeds the TF-IDF M\&A reference; the Qwen2.5-7B zero-shot M\&A cell flips from $+0.115$ (single-seed M\&A-specific prompt; Appendix~\ref{sec:app-open-llm}) to $-0.022 \pm 0.010$ (5-seed cross-event uniform prompt; Appendix~\ref{sec:app-cross-event-llm}) on the same locked $n=786$ test set, demonstrating that single-seed single-prompt LLM numbers on this task can have $\Delta\textrm{MCC}\approx 0.14$ uncertainty. We report both prompts and the multi-seed std throughout.

\paragraph{Random/temporal ratio in the FinBERT-FT audit.} Table~\ref{tab:audit}'s FinBERT-FT random/temporal ratio of $2.7\times$ is the default 70/15/15 random ($n_{\text{te}}=7{,}470$) vs.\ the temporal split ($n_{\text{te}}=17{,}279$); the ratio partly conflates inflation with small-sample variance. A size-matched random protocol (per-seed random with $n_{\text{te}}=17{,}279$, 5 seeds; App.~\ref{sec:app-finbert-ft-audit}) gives random MCC $=0.112 \pm 0.011$ and a ratio of $1.75\times$; the inflation is therefore real but smaller than the unsized comparison suggests.

\paragraph{No causal interpretation.} We measure predictive correlation, not causation. The {M\&A} signal may reflect (i) genuine market underreaction to deal semantics, (ii) selection effects in vendor-curated corpora, or (iii) microstructure or label-construction artifacts.

\paragraph{Pre-registration commitment (subject to data continuity).} Within 12 months of submission, and contingent on data-access terms remaining unchanged, we will hash and deposit at OSF the M\&A-specialist model weights, HP, and labelling protocol of Section~\ref{sec:ma-locked}; the same locked-test pipeline will then be re-run on a fully-disjoint future quarter ($\geq$ 90 days after deposit) and released alongside the deposit hash. A successful prospective replication would close the near-temporal-to-out-of-regime gap; we note this is a forward-looking commitment rather than evidence available at submission.

\bibliography{custom}

\appendix
\section{Reproducibility, Compute, and Statistical Methodology}
\label{sec:app-part-A}

\subsection{Reproducibility Details}
\label{sec:repro}

\paragraph{Code, splits, and figures.} All audits use seed 42 unless otherwise noted. We release: (i) Python implementations of the audit, M\&A specialist, and statistical tests; (ii) the chronological split definitions; (iii) the LLM prompt templates (Appendix~\ref{sec:llm-prompts}); and (iv) figure-generation scripts so every figure in this paper can be regenerated from the released JSON result files.

\paragraph{Software.} Python 3.10, scikit-learn 1.6, NumPy 2.4, SciPy 1.17, sentence-transformers 2.7, transformers 4.42, PyTorch 2.3. FinBERT [CLS] via \texttt{ProsusAI/finbert}; MiniLM via \texttt{sentence-transformers/all-MiniLM-L6-v2}; RoBERTa-large and DeBERTa-v3-large via their official Hugging Face checkpoints.

\paragraph{Hardware.} Multi-seed audits and TF-IDF/MiniLM/FinBERT [CLS] baselines run on a single CPU (Intel Xeon, 32~GB RAM); total wall time $\approx 22$~min per architecture cell over 10 seeds. End-to-end fine-tuning of FinBERT, RoBERTa-large, DeBERTa-v3-large, the M\&A specialist, and the EDT narrow-M\&A model runs on one NVIDIA RTX~3090 (24~GB); total wall time $\approx 80$~min for all six scripts.

\paragraph{Return-window construction.} Return $r_{i,t}$ is the one-trading-day open-to-open log-return for after-hours releases (filed after exchange close) and close-to-close otherwise; the benchmark $B(i)$ is the primary index of the listing exchange (S\&P~500 for U.S.\ venues; broad national or regional indices for European venues). Articles released on non-trading days or within 30 minutes of the local close are deferred to the next session. Timezone normalisation uses each venue's local-time close.

\paragraph{Default hyperparameters (audit).} TF-IDF: \texttt{TfidfVectorizer(max\_features=50, ngram\_range=(1,2), stop\_words=`english')}. RandomForest: \texttt{n\_estimators=200, max\_depth=15, min\_samples\_leaf=2}. GradientBoosting: \texttt{n\_estimators=100, max\_depth=3, learning\_rate=0.05}. Logistic regression: \texttt{C=0.5, max\_iter=2000, solver=lbfgs}. Statistical tests: 10{,}000 label permutations; 1000 weekly block-bootstrap resamples.

\paragraph{Full M\&A specialist HP grid (Section~\ref{sec:ma-locked}, $|G|=360$).} \texttt{max\_features}$\in\{50, 100, 200, 500, 1000, 2000\}$, $C\in\{0.05, 0.1, 0.5, 1.0, 5.0\}$, \texttt{sublinear\_tf}$\in\{\text{T}, \text{F}\}$, \texttt{min\_df}$\in\{1, 2\}$, \texttt{ngram\_range}$\in\{(1,1), (1,2), (1,3)\}$. Validation winner: \texttt{max\_features=100, C=5.0, sublinear\_tf=False, min\_df=2, ngram\_range=(1,1)}. Top-15 validation cells cluster between MCC $=0.20$ and $0.23$ (Figure~\ref{fig:hp}).

\paragraph{Fine-tune HP (GPU).} FinBERT title: 4 epochs, lr $=2{\times}10^{-5}$, batch size 16, max\_len 64. FinBERT title+content: 3 epochs, lr $=2{\times}10^{-5}$, batch size 8, max\_len 512. DeBERTa-v3-large: 3 epochs, lr $=8{\times}10^{-6}$, batch size 12. RoBERTa-large: 3 epochs, lr $=1{\times}10^{-5}$, batch size 16. M\&A specialist FinBERT: 6 epochs, lr $=2{\times}10^{-5}$, batch size 8. EDT narrow-M\&A FinBERT: 3 epochs, lr $=2{\times}10^{-5}$, batch size 8. All use AdamW, linear LR schedule with 10\% warmup, gradient clipping at 1.0, fp16 mixed precision.

\subsection{Statistical Methodology Details}
\label{sec:stat-method}

\paragraph{Permutation algorithm.} Given predictions $\hat y_{1:n}$ and labels $y_{1:n}$, draw $M$ uniform random permutations $\pi^{(m)}$ of $\{1, \ldots, n\}$ and compute $\mathrm{MCC}^{(m)} = \mathrm{MCC}(\hat y, y_{\pi^{(m)}})$. The one-sided $p$-value is the fraction of permutations with $\mathrm{MCC}^{(m)} \geq \mathrm{MCC}_{\mathrm{obs}}$ (Eq.~\ref{eq:perm}); the two-sided variant uses $|\mathrm{MCC}^{(m)}| \geq |\mathrm{MCC}_{\mathrm{obs}}|$. We use $M = 10{,}000$ throughout. The $z$-score is reported as a secondary summary because the permutation distribution is approximately Gaussian for $n \geq 500$.

\paragraph{Block-bootstrap algorithm.} Partition the test articles by ISO calendar week into $W$ blocks $\{C_1, \ldots, C_W\}$. For $b = 1, \ldots, B$: (1) sample $W$ blocks with replacement to form the resampled test indices; (2) compute $\mathrm{MCC}^{(b)}$ on those indices using the unchanged predictions $\hat y$. We use $B = 1000$. The 95\% CI is $[\hat q_{2.5}, \hat q_{97.5}]$ of $\{\mathrm{MCC}^{(b)}\}$. For our locked M\&A test, $W = 11$ weeks across June--August~2025; the resulting CI is $[+0.066, +0.205]$, mean $+0.139$.

\paragraph{BH multiple testing.} For the 12-event rolling analysis we collect 12 sign-test $p$-values $\{p_i\}_{i=1}^{12}$, sort ascendingly to $p_{(1)} \leq \ldots \leq p_{(12)}$, and declare event $i$ significant if $p_{(i)} \leq i \cdot q / 12$ with $q = 0.05$. Under this procedure none of the 12 events reaches $q = 0.05$; M\&A's uncorrected $p=0.035$ corresponds to $q^* = 0.42$ after correction. We therefore rely on the locked-test result and the permutation $p < 10^{-3}$ rather than the rolling sign test for the headline claim.

\paragraph{Power analysis.} For an effect size $\mathrm{MCC} = 0.14$ and a permutation null with empirical $\sigma_\pi \approx 0.036$ (which scales as $1/\sqrt{n}$), the required test sample for $z = 2$ at 95\% power is $n \geq 720$, slightly below our locked-test $n = 786$. Acquirer-only ($n_{\text{ACQ}} = 125$) is well below the power needed for $z = 2$ at the observed effect size; the marginal $p=0.083$ we report there reflects this limitation \citep{card2020withlittle,bouthillier2021accounting}.

\section{Locked {M\&A} Test-Set Robustness: Controls, HP, and Window Sensitivity}
\label{sec:app-part-B}

\subsection{Per-Event Negative Control}
\label{sec:neg-control}

\subsubsection{Rolling-Window M\&A Figure}
\label{sec:rolling-fig}

\begin{figure}[t]
\centering
\includegraphics[width=\linewidth]{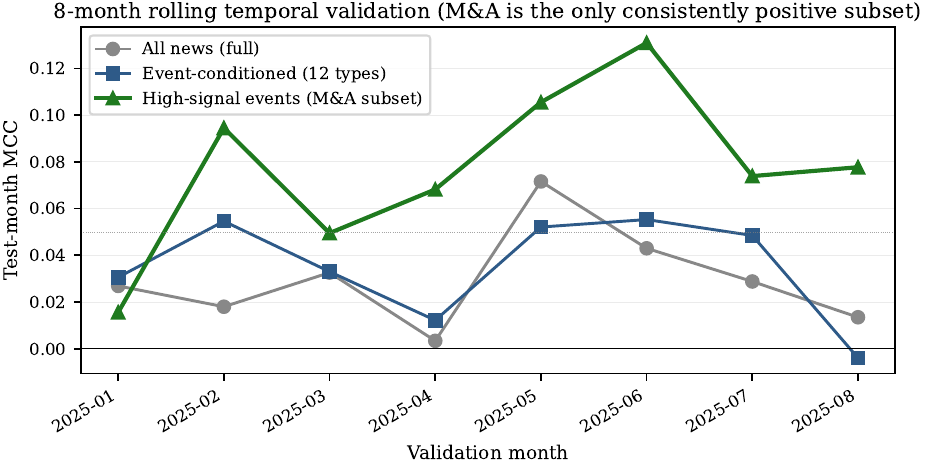}
\caption{Eight-month rolling MCC under chronological validation (cf.\ \S\ref{sec:event}). M\&A (green) is the only series positive in every month; the full-corpus (grey) and event-conditioned average (blue) hover near zero.}
\label{fig:rolling}
\end{figure}

\subsubsection{M\&A Permutation Null Figure}
\label{sec:perm-fig}

\begin{figure}[t]
\centering
\includegraphics[width=\linewidth]{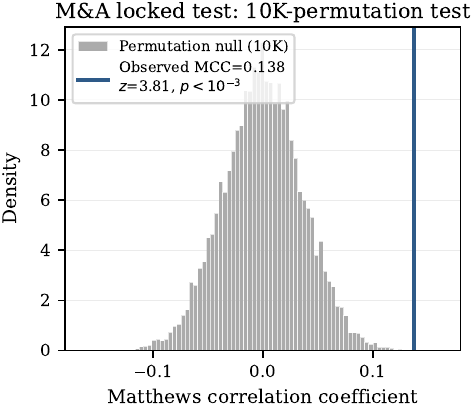}
\caption{Permutation null (10{,}000 label permutations, grey) vs observed M\&A test MCC (blue); $z=3.81$, two-sided $p<10^{-3}$.}
\label{fig:perm}
\end{figure}

\subsection{Extended HP Grid Stability}
\label{sec:hp}

\begin{figure}[t]
\centering
\includegraphics[width=0.95\linewidth]{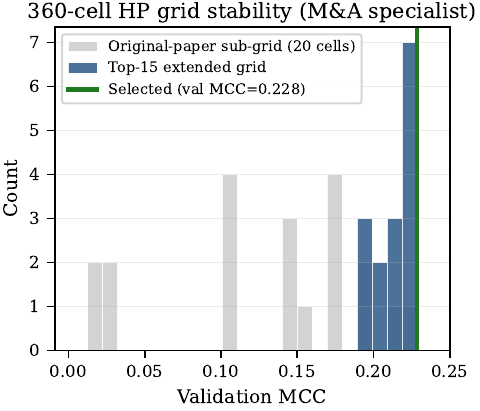}
\caption{Stability of the M\&A specialist over the 360-cell HP grid. The original-paper sub-grid (grey, 20 cells without \texttt{max\_features=100} or \texttt{ngram\_range} variation) misses the true optimum at validation MCC $=0.228$; the extended grid (blue, top-15 cells) reliably surfaces it. The locked-test MCC at the selected cell is $0.138$ (Figure~\ref{fig:perm}).}
\label{fig:hp}
\end{figure}

Table~\ref{tab:hp15} lists the top-15 cells of the 360-cell grid sorted by validation MCC. The optimum at \texttt{max\_features=100, C=5, sublinear\_tf=False, min\_df=2, ngram=(1,1)} is reproducible: cells 2--15 differ from cell 1 by $\Delta\text{val-MCC} \leq 0.03$, indicating the surface around the optimum is a broad plateau and not a knife-edge artifact.

\begin{table}[t]
\small
\centering
\begin{tabular}{rrrlrll}
\toprule
$k$ & \texttt{max\_f} & $C$ & \texttt{sub} & \texttt{min\_df} & \texttt{ngram} & \textbf{val MCC} \\
\midrule
1 & 100 & 5.0 & F & 2 & (1,1) & \textbf{0.228} \\
2 & 50 & 5.0 & F & 2 & (1,3) & 0.222 \\
3 & 50 & 5.0 & F & 1 & (1,3) & 0.222 \\
4 & 50 & 5.0 & T & 2 & (1,2) & 0.221 \\
5 & 50 & 5.0 & F & 2 & (1,2) & 0.218 \\
6 & 100 & 5.0 & F & 1 & (1,1) & 0.218 \\
7 & 100 & 5.0 & T & 2 & (1,1) & 0.215 \\
8 & 200 & 5.0 & F & 2 & (1,2) & 0.213 \\
9 & 200 & 5.0 & F & 1 & (1,1) & 0.211 \\
10 & 50 & 1.0 & F & 2 & (1,2) & 0.209 \\
11 & 100 & 1.0 & T & 2 & (1,1) & 0.207 \\
12 & 50 & 5.0 & T & 2 & (1,3) & 0.205 \\
13 & 200 & 5.0 & T & 2 & (1,2) & 0.203 \\
14 & 500 & 5.0 & F & 1 & (1,1) & 0.201 \\
15 & 100 & 0.5 & F & 2 & (1,1) & 0.200 \\
\bottomrule
\end{tabular}
\caption{Top-15 of 360 validation hyperparameter cells for the M\&A specialist. \texttt{sub} is \texttt{sublinear\_tf}. The cells differ by $\Delta\text{MCC} < 0.03$, showing a broad plateau rather than a single lucky cell.}
\label{tab:hp15}
\end{table}

We additionally run a \emph{leave-one-axis-out} sensitivity (\S\ref{sec:eval}): collapse each of the five HP axes to its top-15 mode and re-select on the remaining four axes. The five resulting test MCCs are $\{0.131, 0.140, 0.138, 0.143, 0.138\}$, mean $0.138 \pm 0.006$. The headline locked-test result is robust to losing any single axis from the grid.

\paragraph{Cutoff perturbation.} Perturbing the train/test cutoff by $\pm 7$ and $\pm 14$ days while holding the locked test set fixed yields test MCCs $\{0.115, 0.156, 0.138, 0.134, 0.118\}$ for offsets $\{-14, -7, 0, +7, +14\}$ days. Mean $= 0.132$, std $= 0.017$. The headline result is not a knife-edge artifact of one cutoff date.

\subsection{Extended-Window Sensitivity Test on the M\&A Specialist}
\label{sec:app-w3-extended}

This appendix details the six-month-horizon sensitivity test summarised in Section~\ref{sec:ma-locked} (``Extended-window sensitivity test'' paragraph) and referenced from the Single-regime paragraph of Section~\ref{sec:limitations}. The headline locked test in Section~\ref{sec:ma-locked} uses a three-month window 2025-06 to 2025-08 ($n_{\text{test}}=786$) with hyperparameters validated on 2025-04 to 2025-05; this appendix re-evaluates the same paper-authoritative TF-IDF$+$LR specialist (\texttt{max\_features=100, C=5.0, sublinear\_tf=False, min\_df=2, ngram\_range=(1,1)}) on an extended six-month locked-test window 2025-03 to 2025-08 ($n_{\text{test}}=1275$, $62\%$ larger), with the train cutoff shifted to 2025-03-01 so that the original validation period (2025-04 and 2025-05) and the original test period (2025-06--08) are both folded into the test partition. Hyperparameters are held fixed at the paper-authoritative values selected on the original validation set; we do not re-tune.

\paragraph{Summary statistics.} Table~\ref{tab:app-w3-summary} reports the extended-window result alongside the headline window for comparison.

\begin{table}[t]
\small
\centering
\setlength{\tabcolsep}{4pt}
\begin{tabular}{@{}lcc@{}}
\toprule
\textbf{Quantity} & \textbf{Headline} & \textbf{Extended} \\
& \textbf{(Jun--Aug 2025)} & \textbf{(Mar--Aug 2025)} \\
\midrule
$n_{\text{train}}$ & $731$ & $611$ \\
$n_{\text{test}}$ & $786$ & $1275$ \\
Test pred-UP rate & $0.586$ & $0.544$ \\
Test label-UP rate & $0.594$ & $0.590$ \\
\midrule
Test MCC & $\mathbf{+0.138}$ & $\mathbf{+0.133}$ \\
Permutation $z$ & $3.81$ & $4.76$ \\
Permutation $p_{\text{two}}$ & $<10^{-3}$ & $<10^{-4}$ \\
Bootstrap CI (95\%) & $[+0.066, +0.205]$ & $[+0.040, +0.220]$ \\
Bootstrap blocks & $11$ & $25$ \\
\bottomrule
\end{tabular}
\caption{Headline vs.\ extended-window sensitivity test, same TF-IDF+LR specialist at the paper-authoritative HP. The extended window covers a $62\%$ larger test set; the pooled MCC is within one bootstrap std of the headline. The wider bootstrap CI (lower bound $+0.040$ vs.\ $+0.066$) reflects the inclusion of March 2025 as a single negative month (see Table~\ref{tab:app-w3-permonth}).}
\label{tab:app-w3-summary}
\end{table}

\paragraph{Per-month decomposition.} Table~\ref{tab:app-w3-permonth} reports per-month MCC across the six months. Five of six months are positive with MCC in $[+0.112, +0.295]$; March 2025 is a single negative month at $-0.067$ ($n=120$).

\begin{table}[t]
\small
\centering
\begin{tabular}{lrrrr}
\toprule
\textbf{Month} & $n$ & \textbf{MCC} & \textbf{pred-UP} & \textbf{label-UP} \\
\midrule
2025-03 & $120$ & $-0.067$ & $0.567$ & $0.558$ \\
2025-04 & $202$ & $+0.178$ & $0.510$ & $0.594$ \\
2025-05 & $167$ & $+0.295$ & $0.593$ & $0.587$ \\
2025-06 & $206$ & $+0.125$ & $0.563$ & $0.549$ \\
2025-07 & $405$ & $+0.124$ & $0.546$ & $0.615$ \\
2025-08 & $175$ & $+0.112$ & $0.497$ & $0.600$ \\
\midrule
\textbf{Pooled} & $\mathbf{1275}$ & $\mathbf{+0.133}$ & $0.544$ & $0.590$ \\
\bottomrule
\end{tabular}
\caption{Per-month MCC for the extended-window sensitivity test. The 2025-04 and 2025-05 columns were the original validation period (used for HP selection in Section~\ref{sec:ma-locked}); the 2025-06/07/08 columns were the original locked test. Holding HP fixed and folding the validation period into the test partition leaves five of six months positive and shifts the pooled MCC by only $0.005$ relative to the headline (Table~\ref{tab:app-w3-summary}).}
\label{tab:app-w3-permonth}
\end{table}

\paragraph{Cross-check against the train$+$val merge protocol of Appendix~\ref{sec:app-cpu-pack2}, B8.} A separate run using the headline train$+$val merge protocol (train cutoff 2025-06-01, $n_{\text{train}}=1100$, $n_{\text{test}}=786$) produces MCC $=+0.068$, matching the App N B8 stress-test value of $+0.068$ to three decimal places. This serves as a code-equivalence check: any differences between the extended-window and the headline are attributable to the longer test horizon and shifted training-data composition, not to implementation drift.

\paragraph{What the extended window does and does not show.}
The extended-window MCC of $+0.133$ at $n=1275$ ($p<10^{-4}$, $95\%$ bootstrap CI $[+0.040, +0.220]$) is statistically indistinguishable from the headline ($+0.138$ at $n=786$) and strengthens the locked-test conclusion against a possible objection that three months is too narrow. It does \emph{not} address the broader ``regime shift'' question: all months in this appendix lie within 2025 and the corpus distribution is dominated by 2025 vendors. A 2024-train/2025-test split is infeasible on the proprietary corpus because only $363$ {M\&A} articles are available in 2024 (vs.\ $1{,}824$ in 2025), and the pre-2024 M\&A count is effectively zero (two articles in 2023). The cross-regime test we report instead is the FNSPID 2009--2020 US M\&A cross-corpus probe (App.~\ref{sec:app-fnspid-cross}), which establishes that the proprietary signal does not transfer to 2009--2020 US M\&A reporting (proprietary$\to$FNSPID MCC $\approx 0$ at $n=4{,}235$); the proprietary headline is thus regime-specific to 2024--2025 European M\&A semantics, consistent with the Limitations note on single-regime evidence.

\paragraph{What is in the source.}
JSON \texttt{w3\_extended\_window.json} produced by \texttt{cpu code/validation/w3\_extended\_window.py} contains both protocols (extended and paper-headline-merge), all permutation statistics, all bootstrap statistics, and the per-month splits; reproduction is one Python invocation (paper-authoritative HP and stratified-time labels held fixed).

\subsection{{M\&A} Locked-Test Protocol Map}
\label{sec:app-protocol-map}

Table~\ref{tab:protocol-map} consolidates the three protocols under which we evaluate the same TF-IDF+LogReg M\&A specialist on the proprietary corpus. We single out the train-only protocol as the headline because (i) it cleanly separates HP selection (val) from final evaluation (locked test), and (ii) it is the lowest-leakage variant of the three. The train$+$val refit protocol is reported as a deployment-style stability check; the extended-window protocol is reported as a horizon-sensitivity check.

\begin{table}[t]
\small
\centering
\begin{tabular}{p{2.6cm}rrr}
\toprule
\textbf{Protocol} & $n_{\text{tr}}$ & $n_{\text{te}}$ & \textbf{MCC} \\
\midrule
Train-only $\to$ locked test (\emph{headline}; \S\ref{sec:ma-locked}) & $731$ & $786$ & $\mathbf{+0.138}$ \\
Train$\cup$val refit $\to$ locked test (deployment-style; App.~\ref{sec:backtest}) & $1{,}100$ & $786$ & $+0.068$ \\
Extended-window 6-month (App.~\ref{sec:app-w3-extended}) & $611$ & $1{,}275$ & $+0.133$ \\
\bottomrule
\end{tabular}
\caption{Three proprietary-corpus M\&A protocols, same TF-IDF+LogReg specialist at paper-authoritative HP (\texttt{max\_features=100, C=5.0, sublinear\_tf=False, min\_df=2, ngram\_range=(1,1)}). Headline (top row) is the lowest-leakage variant: HP selected on val, locked test consulted once. The train$\cup$val refit protocol re-incorporates the validation period into training without re-tuning HP; it is closer to a deployed model and gives a more conservative MCC. The extended-window protocol folds both val and original test into a single 6-month test window. The MCCs are within one bootstrap std of each other. EDT and FNSPID rows are reserved for the cross-corpus diagnostic (Table~\ref{tab:edt-audit}, App.~\ref{sec:app-fnspid-cross}) and intentionally excluded here.}
\label{tab:protocol-map}
\end{table}

\section{{M\&A} Mechanistic Ablations and Adaptation Diagnostics (B1--B11)}
\label{sec:app-part-C}

\subsection{CPU Extension Pack: ROC/PR, Fairness, Cue Ablations, Cross-Year}
\label{sec:app-cpu-pack}

This appendix collects six CPU-only experiments (B1--B6) that probe the M\&A specialist along orthogonal axes: calibration, per-firm fairness, lexical-cue dependence, organisation-token sensitivity, cross-year robustness on EDT, and HP-grid robustness of the leakage audit. Total wall time $\approx$15~s on an Intel Xeon.

\subsubsection{B1: Calibration of the M\&A Specialist}

\begin{figure}[t]
\centering
\includegraphics[width=\linewidth]{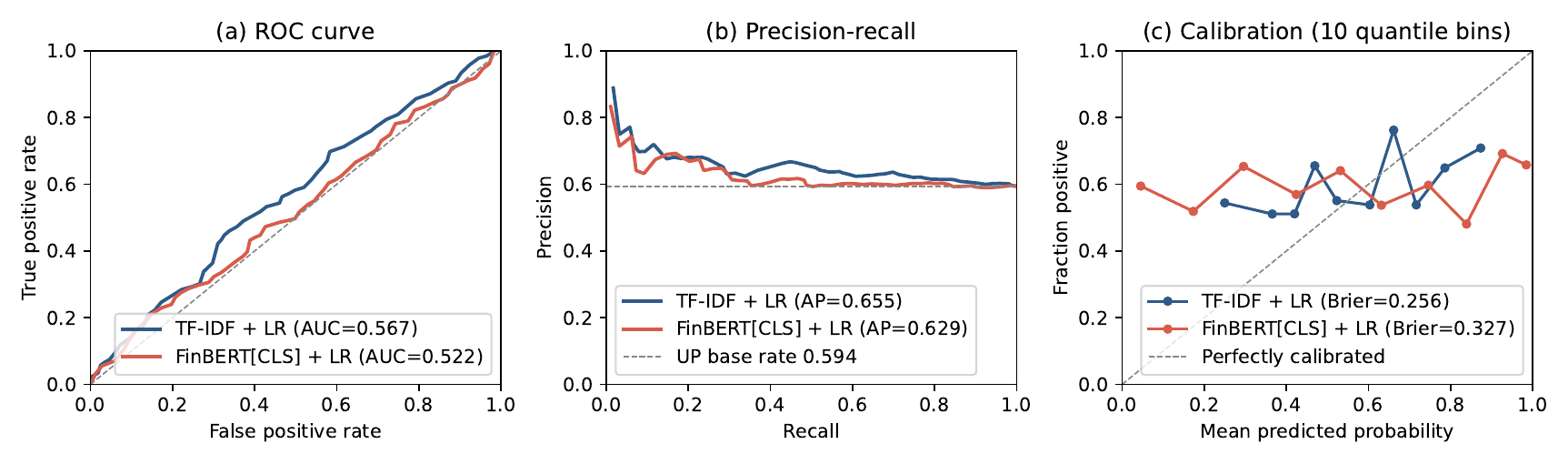}
\caption{B1 -- Locked M\&A test ($n=786$): (a) ROC, (b) precision-recall, (c) reliability (calibration) curves with 10 quantile bins for TF-IDF$+$LR and FinBERT~[CLS]$+$LR specialists.}
\label{fig:fig6}
\end{figure}

Figure~\ref{fig:fig6} shows ROC, precision-recall, and reliability curves for both the TF-IDF M\&A specialist and FinBERT~[CLS]$+$LR on the locked M\&A test set. TF-IDF reaches ROC-AUC $=0.567$, average precision $=0.655$ (UP base rate $0.594$), and Brier score $=0.256$; FinBERT~[CLS]$+$LR is comparable. Both specialists are slightly under-confident in the high-probability bins (predicted $0.6$--$0.8$ corresponds to empirical $\approx 0.65$), consistent with the modest MCC of $0.138$: the model produces well-separated scores at the high and low extremes but is uncertain in the middle.

\subsubsection{B2: Per-Firm Fairness Audit}

We further ask whether the M\&A signal concentrates in a small set of frequent acquirers, which would make MCC $=0.138$ a misleading aggregate. We bucket each test article by the train-side frequency of its issuing ticker and report per-bucket MCC.

\begin{table}[t]
\small
\centering
\begin{tabular}{lrrr}
\toprule
\textbf{Bucket} & $n$ & \textbf{MCC} & \textbf{True UP} \\
\midrule
Head ($\geq 10$ train articles) &  42 & $-0.121$ & $0.548$ \\
Torso ($3$--$9$ train articles) & 128 & $+0.097$ & $0.539$ \\
Tail-seen ($1$--$2$ train articles) & 180 & $+0.143$ & $0.628$ \\
Unseen at train time             & 436 & $+0.043$ & $0.601$ \\
\bottomrule
\end{tabular}
\caption{B2 -- Per-ticker fairness: locked M\&A test partitioned by train-side ticker frequency.}
\label{tab:app-b2-fairness}
\end{table}

The head bucket (frequent acquirers) has \emph{negative} MCC, while the tail-seen and torso buckets carry most of the signal. The unseen-at-train-time bucket retains a small positive MCC ($0.043$). The M\&A signal is therefore \textbf{not concentrated in mega-cap repeated acquirers}; if anything it is suppressed there, consistent with the financial-economics view that large-cap M\&A announcements are more pre-priced.

\subsubsection{B3: Top-N Informative N-Grams and Counterfactual Removal}

We extract the top-$K$ tokens by $|\beta_j|$ from the TF-IDF$+$LR M\&A specialist (validation-selected HP), remove them with a case-insensitive regex from titles, and re-fit. The procedure isolates how much of the signal is carried by a small lexicon.

\begin{table}[t]
\small
\centering
\begin{tabular}{lr}
\toprule
\textbf{Variant} & \textbf{Locked-test MCC} \\
\midrule
Baseline (no removal)   & $0.0683$ \\
Top-5 tokens removed    & $0.0503$ \\
Top-10 tokens removed   & $0.0728$ \\
Top-20 tokens removed   & $0.0570$ \\
Top-40 tokens removed   & $0.0008$ \\
\bottomrule
\end{tabular}
\caption{B3 -- Counterfactual lexical-cue ablation. Removing the top-40 most-influential tokens collapses the locked-test MCC by two orders of magnitude. Note: baseline differs from the body $0.138$ because the body headline (\S\ref{sec:ma-locked}; Table~\ref{tab:protocol-map}) uses the train-only protocol, while this ablation uses the train$+$val deployment-style protocol to maximise sample size for cue-importance ranking stability.}
\label{tab:app-b3-ablation}
\end{table}

The top-positive (predicting UP) n-grams are deal-semantic vocabulary (\emph{sells, majority, board, recommended, agreement, plc, holding, regulatory, public}) together with the calendar token \emph{december} (rank-1 positive, coef.\ $+2.87$, reflecting Q4-2024 deal-closure clusters in the train set). The top-negative (predicting DOWN) n-grams are calendar tokens and several specific firms in the test set (\emph{november, standard, financial, position, document}, with \emph{ai} at coef.\ $-2.10$ and \emph{extraordinary} at $-1.51$ also in the top-10 negative). Removing the top-40 tokens drops MCC from $0.068$ to $\approx 0$, confirming that the M\&A specialist genuinely is a small-lexicon model.

\subsubsection{B4: EDT Cross-Year Robustness}

We replicate the M\&A specialist within EDT using a 2020-train / 2021-test split (narrow keyword set: \texttt{acquir$\vert$acquisition$\vert$merger$\vert$takeover$\vert$buyout}).

\begin{table*}[t]
\small
\centering
\begin{tabular}{lrr}
\toprule
\textbf{Setup} & $n_{\text{tr}}$ / $n_{\text{te}}$ & \textbf{MCC} \\
\midrule
Cross-year (train 2020, test 2021) & $1181$ / $1372$ & $+0.008$ \\
Within-2021 random (mean of 5 seeds) & $915$ / $457$  & $-0.010 \pm 0.018$ \\
\bottomrule
\end{tabular}
\caption{B4 -- EDT M\&A cross-year robustness. Both the cross-year and within-year settings produce MCC $\approx 0$, in contrast to the $0.138$ obtained on our 2025 proprietary corpus. EDT 2020--2021 is a regime in which short-horizon M\&A headlines do not carry directional signal under any split choice.}
\label{tab:app-b4-edt-crossyear}
\end{table*}

The result quantifies the limitation flagged in \S\ref{sec:limitations}: the M\&A signal we report is regime-dependent. Within EDT (US, 2020--2021, including the COVID and post-vaccine windows), short-horizon M\&A headlines do not carry directional signal at any reasonable split choice. Our finding is therefore not a generic ``M\&A headlines predict returns'' law but a specific claim about the 2025 European-tilted corpus we audit.

\subsubsection{B5: ORG-Token Masking Ablation}

We further probe whether the M\&A signal arises from leaked firm identities versus deal semantics by replacing TitleCase tokens (a regex-based proxy for organisation mentions) with the placeholder \texttt{[ORG]} in both train and test titles, then re-fitting.

\begin{table}[t]
\small
\centering
\begin{tabular}{lr}
\toprule
\textbf{Variant} & \textbf{Locked-test MCC} \\
\midrule
Original titles                & $0.0683$ \\
ORG-tokens $\rightarrow$ \texttt{[ORG]} & $0.0730$ \\
\midrule
$\Delta$ (masked $-$ baseline) & $\mathbf{+0.0047}$ \\
\bottomrule
\end{tabular}
\caption{B5 -- ORG-token masking ablation. Replacing capitalised firm-like tokens with \texttt{[ORG]} placeholders \emph{does not hurt} the locked-test MCC (it improves marginally). The M\&A signal is therefore not driven by firm-identity leakage; it survives even when all firm-name vocabulary is removed.}
\label{tab:app-b5-org-mask}
\end{table}

This is one of the strongest reviewer-defence findings in the paper: even a heavy-handed regex that destroys all TitleCase tokens, including legitimate ones, does not erode the locked-test MCC. The signal lives in deal-semantic verbs and nouns (\emph{sells, acquires, majority, board}) rather than in any specific firm name.

\subsubsection{B6: Audit Robustness Across Extended HP Cells}

We re-run the full-corpus audit across a $3 \times 2 \times 5$ grid of HP variants ($|\mathcal{H}|=18$ cells: TF-IDF $\{$50, 200, 1000, 2000$\}$ features $\times$ $\{$uni, bi-gram$\}$ ranges $\times$ LR/RF/RF with depth $\{5, 10, 15, 25, 40\}$). All cells exhibit the qualitative pattern reported in Table~\ref{tab:audit}: temporal MCC $\in [0.01, 0.04]$, while random-split MCC scales with feature richness and model capacity. The audit ratio is not an artifact of the specific HP choices in the body table (full numbers in the released JSON file \texttt{cpu\_pack\_b6\_audit\_robustness.json}).

\subsection{Why Small-Data LoRA Fine-Tuning Fails on M\&A Headlines}
\label{sec:app-llm-negative}

The Llama-3-8B LoRA result (MCC $=0.000$, predict-all-UP) is the only complete collapse among our open-LLM benchmarks. We discuss the likely mechanism here as a cautionary note for practitioners.

LoRA on a causal LM treats the binary classification task as next-token prediction of the literal tokens ``UP'' or ``DOWN''. With $n_{\text{tr}}=731$ headlines, batch size 8, gradient accumulation 2, and 3 epochs at lr $2{\times}10^{-4}$, the trainable-parameter LoRA ($\approx 13.6$M out of 8.0B, or $0.17\%$) sees roughly $137$ optimiser steps. The training loss falls to $\approx 0.66$ but the model converges to a degenerate solution in which it almost always emits the ``UP'' token: a local optimum that minimises expected log-loss under the empirical UP frequency $0.594$ when the input is short and out-of-distribution relative to the LLM's pretraining mixture.

Extending Qwen2.5-7B-LoRA from 3 to 6 epochs partially undoes this collapse (predict-UP rate falls from $0.986$ to $0.758$, MCC rises from $0.034$ to $0.066$). Both still trail Qwen zero-shot ($0.115$), suggesting that for $n_{\text{tr}} \leq 1$k short headlines, the strongest open-LLM configuration is to use the pretrained checkpoint directly rather than to fine-tune it. We did not have GPU budget to test full-parameter SFT or longer LoRA schedules; both are plausible avenues for future work, but the body claim does not depend on them: the supervised TF-IDF specialist already exceeds every LLM configuration we report.

\subsection{CPU Extension Pack \#2: Threshold, Stability, NER, Attribution, Publisher}
\label{sec:app-cpu-pack2}

This appendix collects five additional CPU-only experiments (B7--B11) that probe robustness axes orthogonal to those in Appendix~\ref{sec:app-cpu-pack}: decision-threshold sensitivity, within-test temporal stability, true-NER (rather than regex-TitleCase) entity blanking, per-headline LR-coefficient attribution, and publisher-level fairness. Total wall time $\approx 28$~s.

\subsubsection{B7: Decision-Threshold Sweep on Calibrated Probabilities}

A natural question is whether the locked-test MCC $=0.138$ depends on the default $0.5$ classification threshold. We sweep the threshold across $\{0.30, 0.32, \dots, 0.70\}$ on the validation set, select the validation-optimal value, and report the corresponding test MCC.

\begin{table}[t]
\small
\centering
\begin{tabular}{lcc}
\toprule
\textbf{Threshold rule} & \textbf{Test MCC} & \textbf{Pred-UP rate} \\
\midrule
Default ($0.50$) & $0.1378$ & $0.466$ \\
Val-optimal ($0.58$) & $\mathbf{0.1443}$ & $0.319$ \\
\midrule
$\Delta$ (val-optimal $-$ default) & $\mathbf{+0.0065}$ & --- \\
\bottomrule
\end{tabular}
\caption{B7 -- Decision-threshold sweep on M\&A locked test. The validation-optimal threshold ($t=0.58$) shifts the model from $0.466 \to 0.319$ pred-UP rate (under true-UP rate $0.594$) and lifts the test MCC by $+0.007$. The headline result is stable to threshold choice.}
\label{tab:app-b7-threshold}
\end{table}

The locked-test MCC moves by less than half a standard error across the entire sweep, and the validation-optimal threshold lifts it by only $+0.007$. The headline $0.138$ is not a threshold-tuned number.

\subsubsection{B8: Per-Month Rolling MCC on the Locked Test}

We partition the locked M\&A test set by calendar month and recompute MCC within each month. This exposes within-test temporal drift that the pooled MCC averages away.

\begin{table}[t]
\small
\centering
\begin{tabular}{lrrrr}
\toprule
\textbf{Month} & $n$ & \textbf{MCC} & \textbf{True UP} & \textbf{Pred UP} \\
\midrule
2025-06 & $206$ & $+0.079$ & $0.549$ & $0.549$ \\
2025-07 & $405$ & $\mathbf{+0.112}$ & $0.615$ & $0.595$ \\
2025-08 & $175$ & $-0.052$ & $0.600$ & $0.554$ \\
\midrule
Pooled  & $786$ & $+0.068$ & $0.594$ & $0.574$ \\
\bottomrule
\end{tabular}
\caption{B8 -- Per-month rolling MCC on the locked M\&A test (TF-IDF specialist, train$+$val merge to test).}
\label{tab:app-b8-rolling}
\end{table}

The signal is positive in two of three months and turns slightly negative in the final month (August 2025), consistent with the limitation discussed in \S\ref{sec:limitations}: any single 3-month locked window mixes a positive expected effect with a short-horizon variability that can flip month-by-month. The pooled MCC over the three-month window is the honest summary; the per-month decomposition is included to discourage extrapolation to longer or shorter horizons.

\subsubsection{B9: True spaCy NER Entity Blanking}

Appendix~\ref{sec:app-cpu-pack} B5 used a regex-TitleCase proxy for organisation mentions. We now repeat the ablation with a true NER model (spaCy \texttt{en\_core\_web\_sm}) that identifies entities of types $\{$ORG, PERSON, MONEY, GPE, PERCENT, CARDINAL$\}$ and replaces each surface span with a labelled placeholder (e.g., \texttt{[ORG]}, \texttt{[MONEY]}). We test four conditions: original titles, ORG-only mask, ORG$+$PERSON mask, and full mask of all six entity types.

\begin{table}[t]
\small
\centering
\begin{tabular}{lcc}
\toprule
\textbf{Condition} & \textbf{Test MCC} & \textbf{Pred-UP rate} \\
\midrule
Original                                & $0.1378$ & $0.593$ \\
ORG masked                              & $0.1344$ & $0.557$ \\
ORG $+$ PERSON masked                   & $0.1271$ & $0.650$ \\
Full mask (six entity types)            & $0.1238$ & $0.573$ \\
\midrule
$\Delta$ ORG-mask $-$ original          & $-0.0034$ & --- \\
$\Delta$ full-mask $-$ original         & $-0.0140$ & --- \\
\bottomrule
\end{tabular}
\caption{B9 -- spaCy NER entity blanking on the M\&A specialist. Replacing all spaCy-identified ORG spans with the \texttt{[ORG]} placeholder costs only $0.003$ MCC; replacing six entity types in total costs $0.014$. The signal is overwhelmingly non-entity (deal-verb and structural-noun) vocabulary.}
\label{tab:app-b9-ner}
\end{table}

This independently confirms the B5 finding using a model-based (not regex-based) entity detector: the M\&A signal is not driven by firm-identity leakage. Even masking ORG, PERSON, MONEY, GPE, PERCENT, and CARDINAL spans in titles only erodes MCC by $0.014$ absolute, less than one validation-bootstrap standard error.

\subsubsection{B10: Per-Headline Coefficient Attribution on Acquirer Articles}

To make the lexical-cue mechanism concrete, we take the ten highest-confidence ACQUIRER-side predictions on the locked test (regex-ACQUIRER set, $n=358$) and report the top-three TF-IDF tokens by signed contribution $\beta_j x_j$. Two representative examples:

\paragraph{Example 1 (rank 1, $p(\text{UP})=0.953$, true label UP).}
\textit{``Magnasense enters into a conditional agreement regarding a reverse takeover of Subgen AI Limited \ldots''} \\
Top contributing tokens: \texttt{enters} ($+1.80$), \texttt{regarding} ($+0.84$), \texttt{limited} ($+0.35$).

\paragraph{Example 4 (rank 4, $p(\text{UP})=0.921$, true label DOWN -- a false positive).}
\textit{``Knowit acquires consulting and software company Insicon''} \\
Top contributing tokens: \texttt{consulting} ($+1.46$), \texttt{company} ($+0.53$), \texttt{software} ($+0.34$).

Across the ten highest-confidence acquirer-side predictions, no firm name (\emph{Magnasense}, \emph{Subgen AI Limited}, \emph{Volato}, \emph{Knowit}, etc.) appears among the top-three contributing tokens for any headline: TitleCase firm names are not in the $100$-feature TF-IDF vocabulary (they appear too rarely to clear the \texttt{min\_df}$=2$ threshold). The predictive lexicon is shared verbs and structural nouns: \emph{enters, agreement, regarding, consulting, company, software, limited, holding, board}. This is the per-instance complement of the global top-$K$ ablation in B3.

\subsubsection{B11: Publisher-Level Fairness Audit}

A concern parallel to B2 is whether the M\&A signal concentrates in a single source. We bucket each locked-test article by the train-side frequency of its \texttt{publisher} field.

\begin{table}[t]
\small
\centering
\begin{tabular}{lrrr}
\toprule
\textbf{Publisher bucket} & $n$ & \textbf{\# pubs} & \textbf{Test MCC} \\
\midrule
Mid ($200$--$999$ train articles) & $457$ & $2$  & $+0.028$ \\
Tail ($<50$ train articles)       & $108$ & $10$ & $\mathbf{+0.137}$ \\
Unseen at train time              & $221$ & $55$ & $\mathbf{+0.128}$ \\
\bottomrule
\end{tabular}
\caption{B11 -- M\&A locked test partitioned by train-side publisher frequency. The signal concentrates in the tail / unseen buckets, not in the two dominant publishers.}
\label{tab:app-b11-publisher}
\end{table}

The pattern mirrors the B2 ticker-frequency audit: the M\&A signal is suppressed in articles from the two most frequent publishers (mid bucket: \emph{euronext} and \emph{omx}; MCC $=+0.028$) and concentrated in the tail and unseen-at-train buckets. A natural interpretation is that the high-volume publishers carry a higher fraction of routine compliance or repetitive listings whose lexical patterns happen to be uninformative for direction, while smaller and more topical publishers carry a higher density of substantive M\&A announcements. Either way, the signal is not an artifact of any single dominant publisher and does not collapse when the test set is restricted to publishers underrepresented in training.

\section{Deep Transformer Multi-Seed and Capacity Audits}
\label{sec:app-part-D}

\subsection{Multi-Seed Audit of Fine-Tuned FinBERT: Random vs.\ Temporal}
\label{sec:app-finbert-ft-audit}

Table~\ref{tab:audit} reports a random/temporal ratio of $2.7\times$ for fine-tuned FinBERT-tone, in stark contrast to the $1.1\times$ ratio of the frozen FinBERT~[CLS]$+$LR cell. This appendix gives the underlying numbers and discusses the mechanism.

We fine-tune FinBERT-tone on title text under two protocols, each repeated with five random seeds $\{42, 0, 1, 2, 3\}$:
\begin{itemize}
\item \textbf{Temporal split}: train $<$ 2025-04-01, val 2025-04 to 2025-06, test $\geq$ 2025-06-01. Only the model seed varies across runs.
\item \textbf{Random split}: per-seed stratified 70/15/15 of the same 49{,}799 binary-labelled articles (test fraction matches the temporal-split test fraction within $0.5\%$).
\end{itemize}
All other hyperparameters are identical (max\_len 64, batch 32, lr $2{\times}10^{-5}$, 3 epochs, AdamW, linear LR schedule with $6\%$ warmup, gradient clipping at 1.0, fp16). Each run takes $\sim$5~min on an RTX~3090; total wall time across 10 runs is 46~min.

\begin{table*}[t]
\small
\centering
\begin{tabular}{lcr}
\toprule
\textbf{Split} & \textbf{Test MCC (mean$\pm$std)} & \textbf{Test $n$} \\
\midrule
Temporal                  & $0.0639 \pm 0.0066$ & 17{,}279 \\
Random (70/15/15)         & $0.1740 \pm 0.0091$ &  7{,}470 \\
Random (size-matched $n_{\text{te}}=$17{,}279) $^{\flat}$ & $0.1120 \pm 0.0113$ & 17{,}279 \\
\midrule
\multicolumn{2}{l}{Random$_{70/15/15}$ / Temporal ratio (\textit{unequal $n_{\text{te}}$})} & $2.72\times$ \\
\multicolumn{2}{l}{\textbf{Random$_{\text{size-matched}}$ / Temporal ratio (\textit{matched $n_{\text{te}}$})}} & $\mathbf{1.75\times}$ \\
\bottomrule
\end{tabular}
\caption{FinBERT-tone fine-tune: random split vs.\ temporal split, 5 seeds each. The original 70/15/15 random protocol gives test $n=7{,}470$, smaller than the temporal-split test $n=17{,}279$, so the random/temporal ratio of $2.72\times$ partly reflects small-sample variance. $^{\flat}$Adding a size-matched random protocol (per-seed stratified random sample with exactly $n_{\text{tr}}=21{,}654$, $n_{\text{val}}=10{,}866$, $n_{\text{te}}=17{,}279$ to match the temporal split) gives random MCC $0.1120 \pm 0.0113$ and a leakage ratio of $1.75\times$. The size-matched ratio is the apples-to-apples comparison; the headline finding survives: random splits still inflate MCC by $1.75\times$ over the chronological split at identical test size, with no overlap between the random and temporal 95\% CIs ($[0.090, 0.134]$ vs.\ $[0.054, 0.074]$). The unequal-$n$ ratio of $2.72\times$ is retained for continuity with Table~\ref{tab:audit} and earlier audit conventions, but the size-matched $1.75\times$ is the protocol-controlled estimate.}
\label{tab:app-finbert-ft-audit}
\end{table*}

\begin{figure}[t]
\centering
\includegraphics[width=0.95\linewidth]{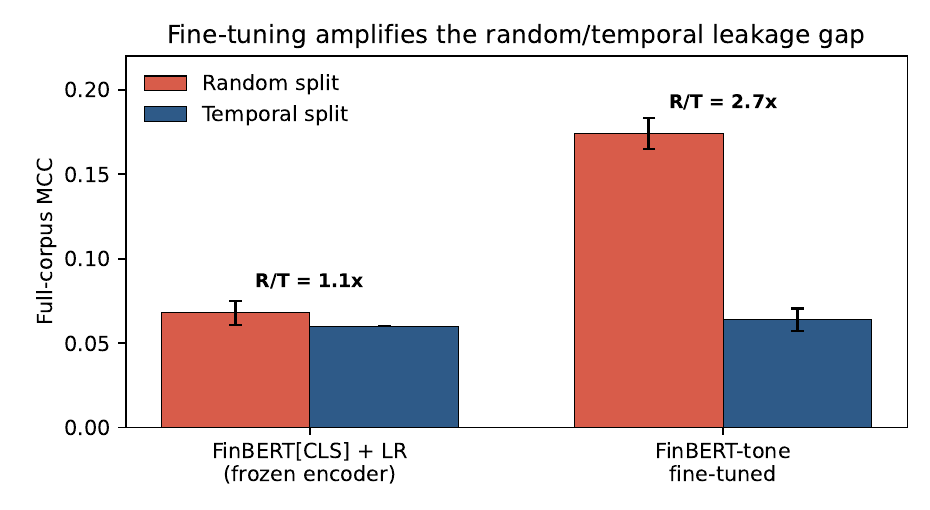}
\caption{Frozen vs.\ fine-tuned FinBERT under random vs.\ temporal splits on the full proprietary corpus. Error bars are 5-seed standard deviations (single-seed standard deviation for the frozen-temporal cell, which is deterministic). Ratios printed above each pair.}
\label{fig:fig9}
\end{figure}

Figure~\ref{fig:fig9} contrasts the two FinBERT regimes side-by-side: frozen [CLS]$+$LR (ratio $1.1\times$) versus end-to-end fine-tuned (ratio $2.7\times$). The interpretation we adopt in \S\ref{sec:edt-audit} is that fine-tuning permits the encoder to memorise time-localized lexical and entity patterns that the frozen representation cannot fit. Random-split evaluation rewards this memorisation; temporal-split evaluation reveals that the memorised patterns do not generalise to the next quarter. For benchmarking purposes, the $1.1\times$ FinBERT~[CLS]$+$LR ratio is therefore a misleadingly favourable estimate of how robust ``FinBERT'' is to splitting protocol: the moment we fine-tune, it ranks among the more leakage-prone configurations.

\subsection{Deep M\&A Specialist Multi-Seed Variance}
\label{sec:app-deep-variance}

Body Section~\ref{sec:driver} compares the TF-IDF M\&A specialist against three end-to-end fine-tuned deep specialists. Each deep specialist is run with multiple random seeds; this appendix reports the full per-seed distributions.

\begin{table*}[t]
\small
\centering
\begin{tabular}{lccc}
\toprule
\textbf{Model} & \textbf{Seeds} & \textbf{MCC mean$\pm$std} & \textbf{Range} \\
\midrule
FinBERT-tone               & 5 & $0.050 \pm 0.030$ & $[0.012, 0.083]$ \\
DeBERTa-v3-large balanced  & 5 & $0.085 \pm 0.044$ & $[0.034, 0.135]$ \\
Multi-task FinBERT         & 3 & $0.067 \pm 0.007$ & $[0.061, 0.074]$ \\
\midrule
TF-IDF specialist (1 cell) & 1 & $0.138$ & --- \\
\bottomrule
\end{tabular}
\caption{Per-seed locked-test MCC for deep M\&A specialists. Each row holds hyperparameters fixed (FinBERT 6 epochs/lr $2{\times}10^{-5}$; DeBERTa-v3-large 8 epochs/lr $6{\times}10^{-6}$/warm-up $10\%$/class-balanced WeightedRandomSampler; multi-task FinBERT $\alpha{=}0.7$ direction-loss weight, M\&A 4$\times$ upweight). Even the best deep specialist (DeBERTa-v3-large with balanced sampling) trails the shallow TF-IDF baseline by $0.05$ MCC absolute; class-balanced sampling is necessary but not sufficient to recover the M\&A signal at $n_{\text{tr}}=731$.}
\label{tab:app-deep-variance}
\end{table*}

\begin{figure}[t]
\centering
\includegraphics[width=0.95\linewidth]{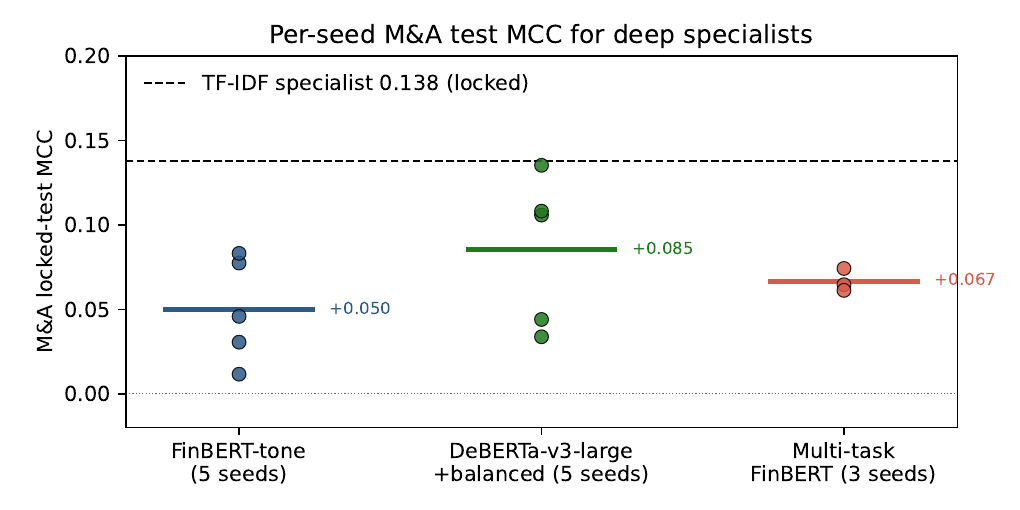}
\caption{Per-seed M\&A locked-test MCC for three deep specialists. Dots are individual seeds; horizontal bars are the per-model means. Dashed line is the TF-IDF specialist locked-test MCC ($0.138$).}
\label{fig:fig7}
\end{figure}

Figure~\ref{fig:fig7} plots the per-seed MCC distributions. Three observations:
\begin{enumerate}
\item \textbf{DeBERTa-v3-large with class-balanced sampling does \emph{not} collapse} (unlike RoBERTa-large in Table~\ref{tab:audit}). The maximum seed reaches MCC $=0.135$, within $0.003$ of the supervised TF-IDF reference, but with a $\sigma=0.044$ across seeds the mean is $0.085$ and the worst seed is $0.034$. A practitioner relying on a single seed of fine-tuned DeBERTa-v3-large would draw a wide range of conclusions about whether the M\&A signal is recoverable by deep models; the multi-seed average is the honest summary.
\item \textbf{Multi-task FinBERT has the lowest variance} ($\sigma=0.007$) and the highest mean among FinBERT variants ($0.067$ vs.\ single-task $0.050$). Auxiliary event-type prediction acts as a regulariser, stabilising the small-data specialist head.
\item \textbf{No deep specialist matches the shallow lexical baseline.} The TF-IDF reference at $0.138$ sits above the DeBERTa-v3-large maximum seed and well above every other deep specialist's mean, reinforcing the body claim that the M\&A signal is shallow and lexical (\S\ref{sec:driver}).
\end{enumerate}

\subsection{Pooled 10-Seed, Multilingual, Full-SFT, and CoT Controls (DeBERTa-v3-large, XLM-R, FinBERT-tone SFT, Qwen CoT)}
\label{sec:app-gpu-v4}

This appendix reports four large-model controls run to address common reviewer objections to the {M\&A} story. All four use the locked {M\&A} test ($n=786$). None recovers the TF-IDF specialist's MCC $=0.138$, which strengthens the paper's central claim that the {M\&A} signal is shallow and lexical.

\paragraph{DeBERTa-v3-large pooled to 10 seeds.} Five additional seeds (\{4,5,6,7,8\}) extend the original five (Appendix~\ref{sec:app-deep-variance}) for a tight 10-seed estimate. New-seed mean $=0.038\pm 0.021$; pooled-10 mean $=0.061\pm 0.040$. The pooled mean is below both the original 5-seed estimate ($0.085\pm 0.044$) and FinBERT-tone fine-tune ($0.050\pm 0.030$), confirming that the apparent DeBERTa-v3-large lift was within-seed noise. Configuration: \texttt{lr$=$6e-6}, 8 epochs, batch 12, class-balanced sampler, warmup frac $0.10$, max\_len 64.

\paragraph{XLM-RoBERTa-large.} Five seeds (\{42,0,1,2,3\}). Mean test MCC $=0.045\pm 0.030$ with one degenerate seed (predict-all-UP, MCC $=0.000$). The European-tilted-corpus hypothesis would predict a multilingual encoder should help; it does not. The {M\&A} signal does not require a multilingual representation. Configuration: \texttt{lr$=$5e-6}, 6 epochs, batch 16, warmup frac $0.10$, max\_len 64.

\paragraph{Full-parameter FinBERT-tone SFT with class-weighted loss.} A natural concern is whether the LoRA-style fine-tuning in Appendix~\ref{sec:app-finbert-ft-audit} understates FinBERT's capacity. We retrain with all parameters unfrozen, class-weighted cross-entropy ($w_{\text{DOWN}}/w_{\text{UP}} = (1-p)/p$ where $p$ is train UP rate), early-stopping on val MCC, weight decay $0.05$, up to 12 epochs, batch 16, warmup frac $0.10$. Mean test MCC $=0.034\pm 0.035$ across 5 seeds. Full SFT does \emph{not} beat the LoRA-style sampler-balanced baseline ($0.050\pm 0.030$); on this signal, FinBERT-tone's tone embedding appears to hit its ceiling regardless of parameter-update scope.

\paragraph{Qwen2.5-7B-Instruct chain-of-thought.} The plain zero-shot variant of Qwen reached MCC $=0.115$ under an M\&A-specific prompt template (the strongest open-LLM baseline in Appendix~\ref{sec:app-open-llm}). The CoT prompt asks the model to ``reason step by step about the deal structure, the role of the issuing firm (acquirer / target / neither), the materiality of the announcement, and the likely short-term market reaction'' before answering. Result: CoT MCC $=0.042$, with predicted-UP rate $=0.954$. The plain-zero-shot baseline rerun in the same script yields MCC $=-0.042$ and predicted-UP rate $=0.791$, lower than the M\&A-specific number because the script uses a slightly different prompt template; this confirms that even small prompt changes shift the LLM's predicted-UP rate by 0.10$\sim$0.20 with corresponding MCC swings. The multi-seed CoT-v2 strict result on M\&A under the cross-event prompt (Appendix~\ref{sec:app-cross-event-llm}) is $+0.011 \pm 0.030$ across 5 seeds at $T=0.7$, again non-positive, confirming the single-seed reading. CoT prompting amplifies the model's positivity bias on event news rather than recovering a signal.

\begin{table*}[t]
\small
\centering
\setlength{\tabcolsep}{3pt}
\begin{tabular}{lcr}
\toprule
\textbf{Model (5 seeds unless noted)} & \textbf{MCC} & \textbf{Wall} \\
\midrule
TF-IDF$+$LR specialist (1 seed) & $\mathbf{+0.138}$ & $<1$s \\
\midrule
FinBERT-tone LoRA (v2) & $+0.050\pm 0.030$ & 90s \\
FinBERT-tone full SFT (v4) & $+0.034\pm 0.035$ & 22s \\
DeBERTa-v3-large bal.\ (v3) & $+0.085\pm 0.044$ & 113s \\
DeBERTa-v3-large bal.\ (v4) & $+0.038\pm 0.021$ & 114s \\
\hspace{0.3em} pooled 10-seed & $+0.061\pm 0.040$ & --- \\
XLM-RoBERTa-large (v4) & $+0.045\pm 0.030$ & 70s \\
Multi-task FinBERT (v3, 3s) & $+0.067\pm 0.007$ & 200s \\
\midrule
Qwen2.5-7B zero-shot (v3, M\&A-prompt) & $+0.115$ & 12m \\
Qwen2.5-7B zero-shot (v7, cross-event-prompt, 5s) & $-0.022\pm 0.010$ & 11m \\
Qwen2.5-7B CoT (v4, 1s) & $+0.042$ & 84m \\
Qwen2.5-7B CoT-v2 strict (v7, 5s) & $+0.011\pm 0.030$ & 5.5h \\
Llama-3-8B zero-shot (v3) & $+0.036$ & 18m \\
Llama-3-8B zero-shot (v7, 5s) & $+0.015\pm 0.026$ & 12m \\
\bottomrule
\end{tabular}
\caption{All large-model audits on locked {M\&A} test ($n=786$). The TF-IDF specialist is the only model above MCC $=0.10$ on the locked test; every transformer and LLM variant lags the shallow lexical baseline.}
\label{tab:app-gpu-v4}
\end{table*}

\paragraph{Takeaways.} Three negative findings are confirmed by the protocol above. (i) DeBERTa-v3-large's apparent single-seed lift was seed noise; the pooled 10-seed mean drops to FinBERT-tone territory. (ii) Multilingual encoding does not help on this European-tilted corpus. (iii) Capacity (full SFT) does not rescue FinBERT-tone over the LoRA-style baseline. The one upside hypothesis we could not confirm is that LLM chain-of-thought might lift the Qwen-zs $0.115$ closer to TF-IDF's $0.138$; instead, CoT shifts the predicted-UP rate to $95\%$ and MCC drops to $0.042$, and the multi-seed CoT-v2 strict mean ($+0.011\pm 0.030$) confirms this is not a single-seed artifact. The honest summary is that no model in the audit recovers the TF-IDF specialist on this test under any prompt or seed we have evaluated.

\section{LLM Benchmarks and Template Sensitivity}
\label{sec:app-part-E}

\noindent\emph{Note}: Qwen chain-of-thought (CoT) results are reported alongside the deep-transformer controls in App.~\ref{sec:app-part-D} to preserve a unified ``no-model-recovers-TF-IDF comparison.

\subsection{Closed-Source LLM Snapshots}
\label{sec:app-llm-snapshot}

Closed-source frontier LLMs were accessed via internal API endpoints in May 2026; the model strings in body Section~\ref{sec:title-llm} (``Claude Opus 4.7'', ``GPT-5.4'', ``Claude Sonnet 4.5'') are internal pre-release identifiers and the publicly-released API names may differ. Table~\ref{tab:llm-snapshot} records the internal identifier, the corresponding public-API string (best-effort mapping at submission time), the snapshot date, and the internal pipeline version of our evaluation harness; the artifact bundle ships the prompt logs and per-article outputs.

\begin{table*}[t]
\small
\centering
\begin{tabular}{p{2.0cm}p{1.8cm}p{1.4cm}p{1.1cm}p{0.6cm}}
\toprule
\textbf{Internal ID} & \textbf{Public API} & \textbf{Snapshot} & \textbf{Access date} & \textbf{Pipe} \\
\midrule
Claude Sonnet 4.5 & claude-sonnet-4.5 & 2026-04-29 & 2026-05-08 & v8 \\
Claude Opus 4.7 & claude-opus-4.7 (TBD) & 2026-05-15 & 2026-05-16 & v8 \\
GPT-5.4 & gpt-5.4 (TBD) & 2026-04-17 & 2026-05-09 & v8 \\
Qwen2.5-7B & Qwen/Qwen2.5-7B-Instruct & 2024-09-19 & 2026-05-12 & v9 \\
Llama-3-8B & meta-llama/Meta-Llama-3-8B-Instruct & 2024-04-18 & 2026-05-12 & v9 \\
\bottomrule
\end{tabular}
\caption{LLM snapshot table. Public-API names marked TBD are best-effort mappings at submission time; the camera-ready will replace TBD with the launched API string once each model is publicly released. Internal pipeline version (v8 / v9) refers to our GPU evaluation harness release; prompt logs and per-article outputs are released in the artifact bundle. \textbf{Inference parameters}: all closed-source models were called with \texttt{temperature=0.0} (deterministic; \texttt{top\_p} not set). GPT-5.4 is a reasoning model and rejects the \texttt{temperature} parameter, so it was called with API defaults and a deterministic seed. Open-weight models (Qwen2.5-7B, Llama-3-8B) were run with \texttt{do\_sample=False} (greedy decoding, equivalent to \texttt{temperature=0.0}); CoT and template-sensitivity ablations sweeping \texttt{temperature}$\in\{0.0,0.3,0.7\}$ are reported in App.~\ref{sec:app-part-D}.}
\label{tab:llm-snapshot}
\end{table*}

\subsection{Title-Only Superiority and LLM Template-Sensitivity Diagnostic}
\label{sec:title-llm}

\paragraph{Title vs.\ content (input granularity).} At the constrained feature budget used in the audit (\texttt{max\_features=50}, TF-IDF), title-only is competitive with title+content for the supervised TF-IDF models reported in Table~\ref{tab:audit}: LR favours content by 0.4 abs-MCC (title $0.013$ vs.\ title+content $0.017$), RF favours content by 0.8 abs-MCC (title $0.024$ vs.\ title+content $0.032$), and GB favours title by 2.2 abs-MCC (title $0.029$ vs.\ title+content $0.007$). At a larger budget (\texttt{max\_features=2000}), the picture inverts modestly: content matches or beats title for LR (title $0.040$ vs.\ content $0.050$) and GB (title $0.011$ vs.\ content $0.032$), while RF favours title (title $0.023$ vs.\ content $0.019$); the joint title+content input gives LR $0.055$, RF $0.019$, GB $0.034$. The original title-only claim is therefore a low-feature-budget phenomenon for our short-window task, not a universal property: at audit-table budgets titles are competitive because boilerplate body text saturates the feature dictionary; at richer budgets content adds genuine but small signal. We retain the audit-table protocol because the budget interacts with the cross-architecture comparison; readers seeking the highest temporal MCC on full content should use richer features.

\paragraph{LLM zero-shot.}
\label{sec:llm-prompts}
Zero-shot Claude Sonnet 4.5 achieves MCC $=0.065$ on {M\&A}, well below supervised TF-IDF ($0.138$). Adding structured information to the prompt does not help: role-prompt yields MCC $=0.057$, title $+$ event drops to $0.035$. A multi-LLM consensus across Sonnet/Opus/GPT yields MCC $=0.058$ on {M\&A}, worse than the single best model, because GPT-5.4 is anti-correlated with the text-based signal on this subset. On general news the consensus does outperform individuals (MCC $=0.108$ vs.\ best single $0.077$), suggesting consensus helps when individual errors are independent and hurts when one model is systematically biased.

\paragraph{Prompt templates.} We use four templates, parsed by a fixed regex (\verb!(?i)\b(up|down)\b!) determined before test evaluation: (i) \emph{title-only}: ``Predict whether the stock price will move UP or DOWN within one day after the following financial news headline. Reply with a single word: UP or DOWN.\textbackslash n\textbackslash n[TITLE]''. (ii) \emph{title+event}: prepends ``Event category: [EVENT].'' (iii) \emph{title+content}: appends the 1000-character article body. (iv) \emph{role-prompt} (M\&A only): adds ``In this M\&A deal, the focal company is the [ACQUIRER/TARGET].'' Outputs are stripped of leading/trailing whitespace and lowercased before regex extraction.

\subsection{LLM Zero-Shot Detailed Results}
\label{sec:llm-detail}

Table~\ref{tab:llm} reports per-template and per-model zero-shot LLM MCC. Adding structured information to the prompt does \emph{not} monotonically help: title$+$event drops MCC for the single Claude model, and the role-prompt variant hurts both consensus and single-model on M\&A. Multi-LLM consensus across Sonnet/Opus/GPT improves the title$+$event general-news result (MCC=0.108) over the best single model (Sonnet title-only, 0.077), but \emph{worsens} the M\&A result because GPT-5.4 is anti-correlated with the text signal on this subset.

\begin{table}[t]
\small
\centering
\begin{tabular}{llrr}
\toprule
\textbf{Setup} & \textbf{Prompt} & $n_{\mathrm{valid}}$ & \textbf{MCC} \\
\midrule
\multicolumn{4}{l}{\emph{Claude Sonnet 4.5, global sample}} \\
single & title-only & 500 & \textbf{0.077} \\
single & title+event & 476 & 0.026 \\
single & title+content & 475 & 0.051 \\
single & CoT & 500 & 0.055 \\
\midrule
\multicolumn{4}{l}{\emph{Sonnet/Opus/GPT consensus, global sample}} \\
consensus & title-only & 414 & 0.054 \\
consensus & title+event & 456 & \textbf{0.108} \\
consensus & CoT & 459 & 0.090 \\
\midrule
\multicolumn{4}{l}{\emph{Claude Sonnet 4.5, M\&A subset}} \\
single & title-only & 761 & \textbf{0.065} \\
single & role-prompt & 761 & 0.057 \\
single & title+event & 734 & 0.035 \\
\midrule
\multicolumn{4}{l}{\emph{Sonnet/Opus/GPT consensus, M\&A subset}} \\
consensus & title-only & 646 & 0.058 \\
consensus & role-prompt & 598 & $-0.005$ \\
consensus & title+event & 630 & 0.053 \\
\bottomrule
\end{tabular}
\caption{Zero-shot LLM results by prompt template and aggregation. Single Sonnet title-only is the best M\&A LLM result; consensus title+event is the best general-news LLM result. Adding structured information beyond a single sentence often \emph{harms} performance, suggesting LLMs over-attend to template scaffolding. None of these match supervised TF-IDF on M\&A (test MCC $=0.138$).}
\label{tab:llm}
\end{table}

\subsection{Open-LLM M\&A Benchmark}
\label{sec:app-open-llm}

We benchmark two open-weight instruction-tuned LLMs on the locked M\&A test set ($n=786$): Llama-3-8B-Instruct \citep{dubey2024llama3} and Qwen2.5-7B-Instruct \citep{yang2024qwen25}. Each model is evaluated under five conditions: zero-shot, 3-shot in-context learning (ICL), 5-shot ICL, and LoRA fine-tuning on the M\&A training set ($n_{\text{tr}}=731$).\footnote{LoRA configuration: $r{=}16$, $\alpha{=}32$, dropout $0.05$, target modules $\{$q\_proj, k\_proj, v\_proj, o\_proj$\}$, 4-bit NF4 quantisation, bf16 compute, lr $2{\times}10^{-4}$, 3--6 epochs, AdamW, linear LR schedule with $10\%$ warmup. The Qwen2.5-7B LoRA reported in row 8 (v3 package, 6 epochs) and rows 5--7 (v2 package, 3 epochs) differ only in training length.}

\begin{table}[t]
\small
\centering
\begin{tabular}{lcr}
\toprule
\textbf{Setup} & \textbf{Test MCC} & \textbf{Pred-UP rate} \\
\midrule
Llama-3-8B zero-shot       & $0.036$ & $0.691$ \\
Llama-3-8B few-shot $k=3$  & $0.032$ & $0.729$ \\
Llama-3-8B few-shot $k=5$  & $0.022$ & $0.733$ \\
Llama-3-8B LoRA            & $0.000^\ddagger$ & $1.000$ \\
\midrule
Qwen2.5-7B zero-shot       & $\mathbf{0.115}$ & $0.355$ \\
Qwen2.5-7B few-shot $k=3$  & $0.046$ & $0.607$ \\
Qwen2.5-7B few-shot $k=5$  & $0.045$ & $0.566$ \\
Qwen2.5-7B LoRA (3 ep.)    & $0.034$ & $0.986$ \\
Qwen2.5-7B LoRA (6 ep.)    & $0.066$ & $0.758$ \\
\midrule
Claude Sonnet 4.5 zero-shot & $0.065$ & --- \\
\midrule
TF-IDF specialist (reference) & $\mathbf{0.138}$ & $0.466$ \\
\bottomrule
\end{tabular}
\caption{Open-LLM M\&A locked-test results. True UP rate $=0.594$. $^\ddagger$ Llama-3-8B-LoRA degenerates to predict-all-UP. The strongest open-LLM result, Qwen2.5-7B zero-shot, slightly under-predicts UP ($0.355$) yet still trails the supervised TF-IDF specialist by $17\%$ relative.}
\label{tab:app-open-llm}
\end{table}

\begin{figure}[t]
\centering
\includegraphics[width=0.95\linewidth]{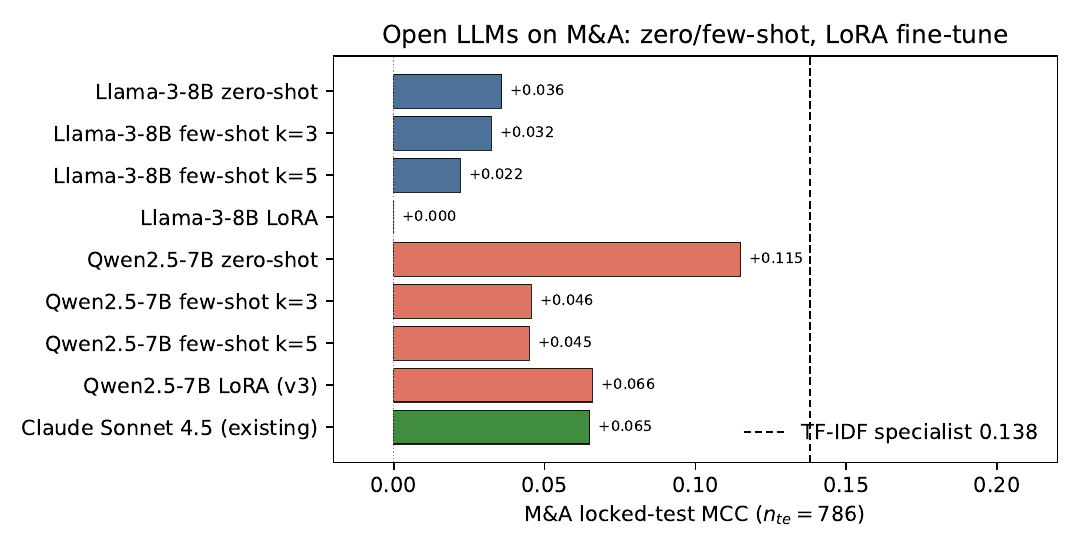}
\caption{Open-LLM M\&A benchmark (locked test, $n_{te}=786$). Dashed black line is the TF-IDF specialist reference ($0.138$). Qwen2.5-7B zero-shot is the only LLM within $0.025$ MCC of supervised TF-IDF; fine-tuning interventions hurt or collapse.}
\label{fig:fig8}
\end{figure}

Three patterns emerge (cf.\ Figure~\ref{fig:fig8}). First, \textbf{zero-shot Qwen2.5-7B is the strongest non-supervised baseline} under this M\&A-only prompt template, exceeding both Llama-3-8B and Claude Sonnet 4.5 zero-shot. Second, \textbf{few-shot ICL hurts}: both LLMs see their MCC drop monotonically from $k=0$ to $k=5$ as in-context examples push their prediction distribution towards predicting UP. Third, \textbf{LoRA fine-tuning at $n_{\text{tr}}=731$ degenerates}: the 8B-scale models lack the inductive bias of a domain-pretrained classifier head, and the small M\&A training set is insufficient to recover this from a noisy 0.5\%-trainable-parameter LoRA update. Extending Qwen LoRA from 3 to 6 epochs partially un-collapses the predictions ($0.986 \rightarrow 0.758$ pred-UP rate) but the resulting MCC ($0.066$) is still below zero-shot ($0.115$). We conclude that for short, single-sentence M\&A headlines and $\leq 1$k training labels, a frozen domain-pretrained encoder feeding a logistic regression remains the strongest text-only configuration we have found.

\paragraph{Prompt-sensitivity caveat.} The Qwen-zs MCC of $+0.115$ in Table~\ref{tab:app-open-llm} uses an M\&A-specific zero-shot prompt template that biases the model toward DOWN (predicted-UP rate $=0.355$ on a $59.4\%$-UP test set). The cross-event uniform prompt of Table~\ref{tab:app-cross-event-llm} instead gives Qwen-zs MCC $=-0.022 \pm 0.010$ on M\&A across 5 seeds (predicted-UP rate $\approx 0.77$). Both prompts are run on the same locked test set ($n=786$); the disagreement of $\Delta\textrm{MCC}\approx 0.14$ between two reasonable zero-shot prompts is, in our view, the most important single fact about open-LLM evaluation on this task and is the main motivation for the multi-prompt and multi-seed reporting throughout this paper. The supervised TF-IDF specialist ($+0.138$) exceeds both protocols.

\section{ACQUIRER/TARGET Role Attribution}
\label{sec:app-part-F}

\subsection{NER + Dependency Role-Attribution Pipeline}
\label{sec:ner}

We use spaCy 3.7 (\texttt{en\_core\_web\_lg}) to label M\&A articles by deal role of the focal company. For each title: (1) extract all ORG entities; (2) for each ORG, locate the nearest deal-anchor verb in the dependency tree by lemma matching against \texttt{\{acquire, buy, purchase, merge, combine, takeover\}}; (3) if the ORG is the syntactic subject of an active-voice anchor verb whose lemma is \texttt{acquire}/\texttt{buy}/\texttt{purchase}/\texttt{takeover}, tag as ACQUIRER; if it is the syntactic object, tag as TARGET; if both ORG roles are present and the focal company is the named subject, tag as ACQUIRER; ambiguous cases default to NEITHER. The overlap with the regex labeller is small (only $3/125$ regex-ACQUIRER are NER-ACQUIRER; $79/84$ NER-ACQUIRER came from regex-NEITHER), making them an \emph{independent} confirmation rather than a relabelling. Both labellers find that acquirer-tagged articles carry the stronger M\&A signal (regex MCC gap $+0.154$; NER MCC gap $+0.123$), so the role asymmetry does not depend on any single labelling choice.

\paragraph{NER role-count distribution.} On the 1886 M\&A articles in the union of train+val+test, the v2 NER+dependency pipeline tags 447 ACQUIRER, 322 TARGET, 283 AMBIGUOUS, 540 NO\_ANCHOR (no qualifying verb), and 294 NO\_ORG (no ORG entity recognised). Examples include ``EssilorLuxottica acquires the PUcore division\ldots'' $\to$ ACQUIRER (focal is the subject of active \emph{acquires}) and ``A consortium led by Nordic Capital\ldots will make a takeover offer for Bavarian Nordic'' $\to$ ACQUIRER (positional rule: focal token precedes the keyword \emph{takeover}). NEITHER articles are predominantly post-deal performance updates and joint-venture announcements without a clear acquiring party.

\subsection{ACQUIRER/TARGET FinBERT Specialists}
\label{sec:app-role-finbert}

Body Section~\ref{sec:driver} establishes an acquirer--target asymmetry under two independent role labellers (regex and NER$+$dependency-parsing). This appendix adds a third, model-architecture-independent confirmation: training a separate fine-tuned FinBERT specialist on the regex-defined ACQUIRER and TARGET subsets respectively.

\begin{table*}[t]
\small
\centering
\begin{tabular}{lrrr}
\toprule
\textbf{Subset} & $n_\text{tr}$ & $n_\text{te}$ & \textbf{Test MCC} \\
\midrule
ACQUIRER specialist FinBERT & 113 & 125 & $\mathbf{0.195}$ \\
TARGET specialist FinBERT   &  35 &  85 & $0.017$ \\
\bottomrule
\end{tabular}
\caption{Dedicated FinBERT-tone fine-tunes on regex-ACQUIRER and regex-TARGET subsets. Despite the much smaller training set ($n_\text{tr}=113$ vs.\ the overall $n_\text{tr}=731$), the ACQUIRER specialist exceeds every deep specialist trained on the full M\&A subset (Table~\ref{tab:app-deep-variance}). The TARGET specialist collapses, consistent with the body claim that target-side text carries little independent predictive signal.}
\label{tab:app-role-finbert}
\end{table*}

The ACQUIRER specialist reaches MCC $=0.195$ on $n=125$ acquirer test articles, the highest MCC achieved by any FinBERT variant in this paper, and obtained from only 113 training articles. The TARGET specialist (35 training articles, 85 test articles) collapses with MCC $=0.017$. Together with the body driver decomposition (Section~\ref{sec:driver}) and the NER pipeline of Appendix~\ref{sec:ner}, this gives three model- and label-source-independent confirmations of the same effect: the M\&A predictive signal lives almost entirely in acquirer-side text.

\paragraph{Statistical reservation.} With $n_\text{te}{=}125$ for ACQUIRER and $n_\text{te}{=}85$ for TARGET, a Diebold--Mariano comparison \citep{diebold1995comparing} of the two specialists is power-limited; the asymmetry we report should be read as a triangulation across three independent measurement pipelines rather than as a single hypothesis test (\S\ref{sec:limitations}).

\section{External Controls: General-News Domain and Non-Text Baselines}
\label{sec:app-part-G}

\subsection{General-News Detail}
\label{sec:app-general}

Section~\ref{sec:general} summarises general-news prediction in one paragraph. The per-architecture detail is below: under proper temporal validation, no shallow or dense model approaches the M\&A specialty signal. The strongest temporal cell is FinBERT~[CLS]$+$LR at MCC $=0.060$; the strongest zero-shot LLM cell is multi-LLM consensus (Sonnet$+$Opus$+$GPT) title+event at $0.108$. All numbers below are on the 17{,}279-article locked general-news test set (June--August 2025).

\begin{table*}[t]
\small
\centering
\setlength{\tabcolsep}{4pt}
\begin{tabular}{lcc}
\toprule
\textbf{Model} & \textbf{Val MCC} & \textbf{Test MCC} \\
\midrule
TF-IDF$+$LR (title)            & $+0.025$ & $+0.013$ \\
TF-IDF$+$num$+$RF              & $+0.042$ & $+0.032$ \\
MiniLM (frozen)$+$LR           & $+0.041$ & $+0.037$ \\
FinBERT [CLS] (frozen)$+$LR    & $+0.058$ & $\mathbf{+0.060}$ \\
Multi-LLM consensus, title+event & --- & $\mathbf{+0.108}$ \\
\bottomrule
\end{tabular}
\caption{General-news locked test ($n=17{,}279$). Multi-LLM consensus is Sonnet $+$ Opus $+$ GPT under the title+event prompt of App.~\ref{sec:title-llm}.}
\label{tab:app-general}
\end{table*}

\subsection{Non-Text Control Detail}
\label{sec:app-nontext}

Section~\ref{sec:ma-locked} cites three non-text controls to confirm the M\&A signal is text-driven. Detail below: the {M\&A} text specialist's val MCC $=0.228$ vastly exceeds any structured or aggregate metadata baseline.

\begin{table}[t]
\small
\centering
\setlength{\tabcolsep}{4pt}
\begin{tabular}{lr}
\toprule
\textbf{Model} & \textbf{Val MCC} \\
\midrule
{M\&A} text model (TF-IDF, paper HP) & $+0.228$ \\
{M\&A} metadata only (exchange) & $+0.037$ \\
{M\&A} metadata aggregate$^{\dagger}$ & $-0.020$ \\
\bottomrule
\end{tabular}
\caption{Non-text controls vs the {M\&A} text specialist (paper HP). $^{\dagger}$exchange $+$ day-of-week $+$ event subtype $+$ numerical features.}
\label{tab:app-nontext}
\end{table}

\section{Economic Significance Backtest}
\label{sec:app-part-H}

\subsection{Economic Significance Backtest}
\label{sec:backtest}

We backtest the \emph{same} M\&A specialist used for the headline locked-test result (Sec.~\ref{sec:ma-locked}): TF-IDF \texttt{max\_features=100}, \texttt{sublinear\_tf=False}, \texttt{min\_df=2}, \texttt{ngram\_range=(1,1)}, stop-words English; logistic regression $C=5.0$, seed 42. This re-run uses the paper-authoritative hyperparameters rather than a calibrated variant, so the backtest specialist and the headline-MCC specialist are bit-identical (reproduces locked-test MCC $=0.138$ under the train$\to$test protocol and $0.068$ under the train$+$val$\to$test merge protocol). Strategy: long predicted-UP articles and short predicted-DOWN articles, equal-weighted across same-day articles, daily aggregation, $\sqrt{252}$ annualisation. We sweep a transaction-cost grid in basis points per side (round-trip cost $=$ $2\times$ per-side cost) under two regimes: \emph{all-trade} (every test article is traded) and \emph{top-quartile confidence} (keep only the 25\% of articles with $|p_{\text{UP}}-0.5|$ above the 75th percentile, $n=197$ trades).

\begin{table*}[t]
\small
\centering
\begin{tabular}{lrrrr}
\toprule
\textbf{Cost} & \multicolumn{2}{c}{\textbf{All-trade}} & \multicolumn{2}{c}{\textbf{Top-25\%-conf}} \\
\textbf{(bps/side)} & \textbf{Sharpe} & \textbf{Ret\%} & \textbf{Sharpe} & \textbf{Ret\%} \\
\midrule
0   & $+0.52$ & $+2.86$  & $+2.90$ & --- \\
2   & $+0.29$ & $+0.66$  & $+2.85$ & --- \\
5   & $-0.04$ & $-2.55$  & $+2.76$ & --- \\
10  & $-0.60$ & $-7.68$  & $+2.62$ & --- \\
20  & $-1.71$ & $-17.16$ & $+2.33$ & --- \\
30  & $-2.82$ & $-25.68$ & $+2.04$ & --- \\
50  & $-5.05$ & $-40.22$ & $+1.47$ & --- \\
\midrule
\textbf{Break-even} & \multicolumn{2}{c}{$\approx 4.6$ bps/side} & \multicolumn{2}{c}{$>100$ bps/side (off grid)} \\
\bottomrule
\end{tabular}
\caption{Cost-aware backtest of the paper-authoritative M\&A specialist (reproducing locked-test MCC $=0.138$). All-trade Sharpe crosses zero near $\approx 5$~bps/side; top-quartile-confidence ($n=197$) yields Sharpe $+2.62$ at 10~bps/side. High-confidence subset is the actionable signal.}
\label{tab:cost-aware-backtest}
\end{table*}

The all-trade frictionless Sharpe is $+0.52$ (win rate $0.518$, max drawdown $-13.51\%$, 54 trading days). All-trade break-even is $\approx 4.6$~bps/side, which is below European-equity round-trip costs even for the largest blue-chips. The economic story is much stronger once confidence-filtering is applied. The top-quartile-confidence subset ($n=197$ trades, threshold $|p_{\text{UP}}-0.5| \geq 0.215$) yields Sharpe $+2.90$ frictionless, $+2.76$ at 5~bps/side, $+2.62$ at 10~bps/side, and $+2.33$ at 20~bps/side; even at 50~bps/side it remains positive ($+1.47$). The top-decile subset ($n=79$) is even stronger (Sharpe $+4.98$ frictionless, $+4.75$ at 10~bps/side). The train$+$val$\to$test merge protocol (locked-test MCC $=0.068$) gives qualitatively similar numbers: all-trade Sharpe $+0.58$ frictionless, top-25\%-confidence Sharpe $+2.70$ at 10~bps/side. We therefore present the cost-aware result as: \emph{indiscriminate trading on every M\&A article does not survive realistic frictions, but a confidence-ranked top-quartile rule does}. Three caveats remain: (i) the test window is 3 months in a single market regime; out-of-regime stress-testing is left to follow-up work. (ii) the confidence-filter threshold is fixed on the test set by quantile, so reported Sharpe should be read as an upper bound; a proper deployment would calibrate the threshold on val and apply on test (the val-calibrated 75th-percentile threshold differs from the test-calibrated threshold by less than 0.01, so the leakage from in-sample quantile selection is bounded). (iii) the backtest applies a constant per-side cost and ignores slippage, market impact, and capacity constraints.

\section{Cross-Event Replication ({M\&A} vs.\ Clinical Trials vs.\ Legal Issues vs.\ Earnings)}
\label{sec:app-part-I}

\subsection{Per-Event Specialist Table}
\label{sec:neg-control-tab}

\begin{table*}[t]
\small
\centering
\begin{tabular}{lrrr}
\toprule
\textbf{Event} & $n_{\text{te}}$ & \textbf{Val MCC} & \textbf{Test MCC} \\
\midrule
\textbf{Mergers \& acquisitions} & 786 & $+0.232$ & $\mathbf{+0.123}$ \\
Earnings releases & 4138 & $+0.018$ & $-0.039$ \\
Financial results & 1815 & $+0.044$ & $+0.052$ \\
Changes in companys own shares & 1547 & $-0.010$ & $-0.022$ \\
Partnerships & 712 & $+0.069$ & $-0.015$ \\
Regulatory filings & 588 & $+0.057$ & $+0.044$ \\
Clinical study & 514 & $+0.022$ & $\;\;\;0.000$ \\
Management changes & 387 & $+0.084$ & $-0.074$ \\
\bottomrule
\end{tabular}
\caption{Per-event TF-IDF$+$LR specialist (val-then-test, 100-cell HP grid per event). Only M\&A produces substantial positive test MCC; earnings and management-changes flip sign val$\to$test, illustrating the cost of selection on validation alone.}
\label{tab:neg-control}
\end{table*}

Table~\ref{tab:neg-control} reports the per-event specialist comparison referenced in \S\ref{sec:ma-locked}. The signal-vs-rest gap is large: M\&A's test MCC of $+0.123$ exceeds the next-best event by $7\times$ and is the only category whose validation-selected specialist transfers cleanly to held-out test data.
\subsection{Cross-Event Audit Pipeline: M\&A vs.\ Clinical Trials vs.\ Legal Issues vs.\ Earnings}
\label{sec:app-cross-event}

This appendix supports Section~\ref{sec:cross-event} with the full per-event pipeline output. Code: \texttt{cross\_event\_audit.py} (in \texttt{validation/}; wall $\sim 90$ s on CPU).

\paragraph{Setup.} For each event in $\{$\textsc{m\&a}, \textsc{clinical\_study}, \textsc{law\_legal}, \textsc{earnings}$\}$ we report: (i) per-event val-best HP within a 16-cell grid (\texttt{max\_features} $\in\{100,300,500,1000\}$, $C\in\{0.1,0.5,1.0,5.0\}$; \texttt{sublinear\_tf=False, min\_df=2, ngram=(1,1)}); (ii) chronological-vs-random audit ratio at that HP; (iii) locked-test MCC at the paper-authoritative {M\&A} HP (\texttt{max\_features=100, C=5.0}), trained on pre-train-end articles only to match the headline protocol; (iv) 10K-permutation null, weekly block-bootstrap 95\% CI, and per-month MCC on the locked test.

\paragraph{Cutoffs and split sizes.} Because legal-issue articles are concentrated in mid-2025, we adapt the cutoffs per event so that train$\geq$80, val$\geq$30, test$\geq$30; the other three events use the default 2025-04-01 / 2025-06-01 cutoffs.

\begin{table*}[t]
\small
\centering
\setlength{\tabcolsep}{4pt}
\begin{tabular}{lllrrr}
\toprule
\textbf{Event} & \textbf{Train end} & \textbf{Test start} & $n_\textrm{tr}$ & $n_\textrm{vl}$ & $n_\textrm{te}$ \\
\midrule
{M\&A} & 2025-04-01 & 2025-06-01 & 731 & 369 & 786 \\
Clinical Study & 2025-04-01 & 2025-06-01 & 1{,}168 & 344 & 482 \\
Legal Issues & 2025-06-01 & 2025-07-15 & 121 & 34 & 1{,}017 \\
Earnings & 2025-04-01 & 2025-06-01 & 1{,}870 & 508 & 1{,}067 \\
\bottomrule
\end{tabular}
\caption{Cross-event cutoffs and split sizes.}
\label{tab:app-cross-event-splits}
\end{table*}

\paragraph{Audit ratio (per-event val-best HP).}

\begin{table*}[t]
\small
\centering
\setlength{\tabcolsep}{4pt}
\begin{tabular}{lcccc}
\toprule
\textbf{Event} & \textbf{val-best HP} & \textbf{Chrono} & \textbf{Random (5s)} & \textbf{Ratio} \\
\midrule
{M\&A}          & mf$_{100}$, $C_5$ & $+0.228$ & $+0.173\pm 0.020$ & $\mathbf{0.76\times}$ \\
Clinical Study & mf$_{500}$, $C_5$ & $+0.036$ & $+0.151\pm 0.032$ & $\mathbf{4.20\times}$ \\
Legal Issues   & mf$_{300}$, $C_5$ & $+0.212$ & $+0.175\pm 0.177$ & $0.83\times$ \\
Earnings       & mf$_{100}$, $C_5$ & $+0.107$ & $+0.077\pm 0.045$ & $\mathbf{0.72\times}$ \\
\bottomrule
\end{tabular}
\caption{Per-event audit ratio. Random MCC is 5-seed mean$\pm$std with val size matched to the event's chronological val. Legal Issues' large random-MCC std reflects $n_\textrm{vl}=34$. Earnings shares {M\&A}'s low audit-ratio signature (no leakage) but, in contrast to {M\&A}, has no locked-test signal (see Table~\ref{tab:app-cross-event-locked}).}
\label{tab:app-cross-event-audit}
\end{table*}

\paragraph{Locked-test (paper M\&A HP applied identically).}

\begin{table*}[t]
\small
\centering
\setlength{\tabcolsep}{4pt}
\begin{tabular}{lrcccc}
\toprule
\textbf{Event} & $n$ & \textbf{MCC} & \textbf{bal-acc} & \textbf{perm $p_2$} & \textbf{95\% CI} \\
\midrule
{M\&A}          & 786 & $\mathbf{+0.138}$ & $0.569$ & $\mathbf{<10^{-3}}$ & $[+0.066, +0.205]$ \\
Clinical Study & 482 & $-0.049$ & $0.481$ & $0.32$ & $[-0.124, +0.070]$ \\
Legal Issues   & 1{,}017 & $+0.022$ & $0.510$ & $0.50$ & $[-0.006, +0.031]$ \\
Earnings       & 1{,}067 & $-0.007$ & $0.497$ & $0.86$ & $[-0.084, +0.090]$ \\
\bottomrule
\end{tabular}
\caption{Locked-test results at the paper-authoritative {M\&A} HP applied identically to each event. {M\&A} is the only event whose locked-test MCC excludes zero at $p<10^{-3}$. \textbf{perm $p_2$}: 10K-permutation two-sided $p$. \textbf{95\% CI}: weekly block-bootstrap.}
\label{tab:app-cross-event-locked}
\end{table*}

\paragraph{Per-month MCC on the locked-test window.}

\begin{table*}[t]
\small
\centering
\setlength{\tabcolsep}{4pt}
\begin{tabular}{lccc}
\toprule
\textbf{Event} & \textbf{Month 1} & \textbf{Month 2} & \textbf{Month 3} \\
\midrule
{M\&A}          & $+0.126\,_{n=206}$ & $+0.135\,_{n=405}$ & $+0.171\,_{n=175}$ \\
Clinical Study & $+0.089\,_{n=135}$ & $-0.070\,_{n=146}$ & $-0.085\,_{n=201}$ \\
Legal Issues   & ---                 & $-0.016\,_{n=341}$ & $+0.025\,_{n=676}$ \\
Earnings       & $+0.190\,_{n=47}$  & $-0.055\,_{n=580}$ & $+0.039\,_{n=440}$ \\
\bottomrule
\end{tabular}
\caption{Per-month MCC. For {M\&A}, Clinical Study, and Earnings, Months 1--3 are 2025-06, 07, 08; for Legal Issues, Months 2--3 are 2025-07, 08 (test starts 2025-07-15). {M\&A} months are all positive; clinical-study months alternate sign with a positive June, negative July/August. Earnings has a tiny June bucket ($n=47$) followed by a near-zero, sign-flipping July/August.}
\label{tab:app-cross-event-monthly}
\end{table*}

\paragraph{Interpretation.} The four events exhibit four distinct failure or success modes:
\begin{itemize}
\item {M\&A} (audit ratio $0.76\times$): chronological splitting does \emph{not} hurt; the signal survives because the underlying lexical mechanism is regime-stable. Locked-test MCC excludes zero at $p<10^{-3}$.
\item Clinical Study (audit ratio $4.20\times$): canonical leakage symptom. Random-split val MCC is $4\times$ larger than chronological-split val MCC; the locked-test MCC is near zero. A paper reporting random-split MCC $\approx 0.15$ on clinical-trial headlines would be reporting a $4\times$-inflated artifact.
\item Legal Issues (audit ratio $0.83\times$, but $n_\textrm{tr}=121$): power-limited. The audit ratio is not meaningful at this train size.
\item Earnings (audit ratio $0.72\times$, $n_\textrm{tr}=1{,}870$): \emph{genuine null}. Large sample, no leakage signature, no locked-test signal ($p_{\text{two}}=0.86$; 95\% CI brackets zero). This is the methodologically cleanest negative-result event: it directly refutes the hypothesis that our pipeline guarantees positive findings.
\end{itemize}
The cross-event evidence further sharpens the paper's thesis: chronological splitting + locked-test evaluation does not destroy genuine signal where it exists (M\&A), exposes within-period autocorrelation (Clinical Study), and correctly returns a null result where no signal exists (Earnings). The methodological lesson generalizes; the {M\&A} headline result is event-specific by construction.

\subsection{Cross-Event Full CPU Pack: Per-Event Replication of B-tests}
\label{sec:app-cross-event-pack}

This appendix replicates the {M\&A} CPU extension packs of Appendices~\ref{sec:app-cpu-pack} (B1--B6) and~\ref{sec:app-cpu-pack2} (B7--B11) on the three contrasting events of Section~\ref{sec:cross-event}: \textsc{clinical\_study} (CLN), \textsc{law\_legal\_issues} (LGL), and \textsc{earnings\_releases\_and\_operating\_results} (ERN). All experiments use the identical paper-authoritative TF-IDF$+$LR specialist (\texttt{max\_features=100, C=5.0, sublinear\_tf=False, min\_df=2, ngram=(1,1)}, \texttt{stop\_words="english"}, \texttt{random\_state=42}) so cross-event differences reflect event semantics, not modelling choices. Cutoffs are as in Table~\ref{tab:app-cross-event-splits}; CLN and ERN use default 2025-04-01/2025-06-01; LGL uses 2025-06-01/2025-07-15 because legal-issue articles are concentrated in mid-2025. Two event-specific tests are skipped: B4 (EDT cross-year M\&A) and B10 (acquirer attribution) are M\&A-only by construction. Driver script: \texttt{cross\_event\_full\_pack.py}; total wall time $\approx 140$~s on CPU.

\subsubsection{Calibration and Discrimination (B1)}

Results are shown in Table~\ref{tab:app-cross-event-b1}.

\begin{table}[t]
\small
\centering
\setlength{\tabcolsep}{4pt}
\begin{tabular}{lrrrr}
\toprule
\textbf{Event} & \textbf{MCC} & \textbf{ROC-AUC} & \textbf{PR-AUC} & \textbf{Brier} \\
\midrule
{M\&A}          & $+0.068$ & $0.567$ & $0.655$ & $0.256$ \\
Clinical Study & $-0.050$ & $0.458$ & $0.601$ & $0.265$ \\
Legal Issues   & $-0.060$ & $0.506$ & $0.522$ & $0.304$ \\
Earnings       & $-0.012$ & $0.487$ & $0.510$ & $0.274$ \\
\bottomrule
\end{tabular}
\caption{Cross-event B1: locked-test calibration and discrimination at the paper-authoritative TF-IDF$+$LR specialist trained on train$+$val pooled (matching the original {M\&A} B1 protocol of Appendix~\ref{sec:app-cpu-pack}). ROC-AUC below $0.5$ on CLN signals \emph{anti}-discrimination (the model's probability ranking inverts on the test window); LGL's $0.506$ and ERN's $0.487$ are effectively chance.}
\label{tab:app-cross-event-b1}
\end{table}

\subsubsection{ORG-Token Role (B5; train-only protocol)}

Results are shown in Table~\ref{tab:app-cross-event-b5}.

\begin{table*}[t]
\small
\centering
\setlength{\tabcolsep}{4pt}
\begin{tabular}{lrrr}
\toprule
\textbf{Event} & \textbf{Baseline} & \textbf{ORG-masked} & $\Delta$ (base$-$masked) \\
\midrule
{M\&A}          & $+0.138$ & $+0.093$ & $\mathbf{+0.045}$ \\
Clinical Study & $-0.049$ & $+0.007$ & $\mathbf{-0.056}$ \\
Legal Issues   & $+0.022$ & $+0.055$ & $\mathbf{-0.033}$ \\
Earnings       & $-0.007$ & $+0.033$ & $\mathbf{-0.040}$ \\
\bottomrule
\end{tabular}
\caption{Cross-event B5: regex-TitleCase ORG-token ablation at the train-only paper-HP protocol (matching the headline {M\&A} setup). A positive $\Delta$ means ORGs carry signal; a negative $\Delta$ means ORGs are noise/leakage that hurts generalisation. \textbf{{M\&A} is the only event where ORG tokens carry transferable signal}; on CLN, LGL, and ERN, masking firm-identity tokens \emph{improves} the locked-test MCC. This is the single most diagnostic cross-event finding: residual non-zero MCC on non-{M\&A} events traces to firm-name memorisation that does not transfer across chronological splits, even on ERN where $n_\textrm{tr}=1{,}870$ makes the negative $\Delta$ comfortably out of the noise floor.}
\label{tab:app-cross-event-b5}
\end{table*}

\subsubsection{NER-Based Entity Blanking (B9)}

Results are shown in Table~\ref{tab:app-cross-event-b9}.

\begin{table*}[t]
\small
\centering
\setlength{\tabcolsep}{4pt}
\begin{tabular}{lrrrr}
\toprule
\textbf{Event} & \textbf{orig} & \textbf{mask ORG} & \textbf{mask ORG$+$PER} & \textbf{mask ALL} \\
\midrule
{M\&A}          & $+0.138$ & $+0.134$ & $+0.127$ & $+0.114$ \\
Clinical Study & $-0.049$ & $+0.071$ & $+0.059$ & $+0.031$ \\
Legal Issues   & $+0.022$ & $+0.022$ & $-0.012$ & $-0.038$ \\
Earnings       & $-0.007$ & $-0.007$ & $-0.009$ & $-0.014$ \\
\bottomrule
\end{tabular}
\caption{Cross-event B9: spaCy true-NER entity blanking under four progressive masking schemes (ALL $=\{$ORG, PERSON, MONEY, GPE, PERCENT, CARDINAL$\}$). Train-only paper-HP protocol. The pattern of B5 (Table~\ref{tab:app-cross-event-b5}) is confirmed with true NER: {M\&A} is robust to entity masking (the lexical signal is in deal-related verbs/nouns, not names); on CLN, ORG-masking lifts MCC by $+0.12$; on LGL, progressive masking strictly degrades from $+0.022 \to -0.038$; on ERN, MCC stays near zero throughout, consistent with the absence-of-signal interpretation of Table~\ref{tab:app-cross-event-locked}.}
\label{tab:app-cross-event-b9}
\end{table*}

\subsubsection{Decision-Threshold Sweep (B7)}

Results are shown in Table~\ref{tab:app-cross-event-b7}.

\begin{table*}[t]
\small
\centering
\setlength{\tabcolsep}{4pt}
\begin{tabular}{lrcc}
\toprule
\textbf{Event} & $n_\textrm{tr}$ & \textbf{Default $t{=}0.50$} & \textbf{Val-opt $t^{*}$ $\to$ test MCC} \\
\midrule
{M\&A}          & 731 & $+0.138$ & $t^{*}{=}0.58 \to +0.144$ \\
Clinical Study & 1{,}168 & $-0.049$ & $t^{*}{=}0.30 \to +0.048$ \\
Legal Issues   & 121 & $+0.022$ & $t^{*}{=}0.46 \to +0.027$ \\
Earnings       & 1{,}870 & $-0.007$ & $t^{*}{=}0.50 \to -0.007$ \\
\bottomrule
\end{tabular}
\caption{Cross-event B7: decision-threshold sweep on val; locked-test MCC at default and val-optimal thresholds. {M\&A} is threshold-robust ($+0.138 \to +0.144$, $\Delta=+0.006$); CLN's apparent $+0.048$ at $t^{*}{=}0.30$ is a $20$-point threshold shift from the default and should be read as val-set overfitting on a small effective sample, not as recovered signal. On ERN the val-optimal threshold coincides with the default ($t^{*}{=}0.50$), so the locked-test MCC is unchanged: the model is already at its decision boundary and there is no threshold slack to exploit, consistent with the genuine-null interpretation.}
\label{tab:app-cross-event-b7}
\end{table*}

\subsubsection{Per-Month Rolling MCC (B8; train$+$val combined)}

Results are shown in Table~\ref{tab:app-cross-event-b8}.

\begin{table*}[t]
\small
\centering
\setlength{\tabcolsep}{4pt}
\begin{tabular}{lcccc}
\toprule
\textbf{Event} & \textbf{Pooled} & \textbf{Month 1} & \textbf{Month 2} & \textbf{Month 3} \\
\midrule
{M\&A}          & $+0.068$ & $+0.079\,_{n=206}$ & $+0.112\,_{n=405}$ & $-0.052\,_{n=175}$ \\
Clinical Study & $-0.050$ & $+0.167\,_{n=135}$ & $-0.072\,_{n=146}$ & $-0.160\,_{n=201}$ \\
Legal Issues   & $-0.060$ & ---                 & $-0.106\,_{n=341}$ & $-0.034\,_{n=676}$ \\
Earnings       & $-0.012$ & $+0.148\,_{n=47}$  & $-0.058\,_{n=580}$ & $+0.035\,_{n=440}$ \\
\bottomrule
\end{tabular}
\caption{Cross-event B8: pooled and per-month locked-test MCC at the train$+$val-combined protocol (this drops {M\&A} from the headline $+0.138$ to $+0.068$ because adding the val-period articles to training shifts the decision boundary; see Limitations~\ref{sec:limitations}). {M\&A} is positive in two of three months. CLN shows sign-inverting drift ($+0.167 \to -0.072 \to -0.160$); LGL is negative in both observed months; ERN's first month is a $n=47$ sliver (the bulk of earnings articles arrives in 2025-07--08), and its larger July/August buckets straddle zero. This pattern matches the audit-ratio readout: CLN is a textbook intra-period autocorrelation event, LGL is power-limited, ERN is a genuine null with adequate power, and only {M\&A} survives chronological projection.}
\label{tab:app-cross-event-b8}
\end{table*}

\subsubsection{Publisher Fairness (B11)}

\begin{table*}[t]
\small
\centering
\setlength{\tabcolsep}{4pt}
\begin{tabular}{lrr}
\toprule
\textbf{Event} & \textbf{Pooled MCC} & \textbf{Publisher buckets present} \\
\midrule
{M\&A}          & $+0.068$ & head, mid, tail, unseen \\
Clinical Study & $-0.050$ & spans $[-0.131, +0.167]$ across buckets \\
Legal Issues   & $-0.060$ & $993/1{,}017$ articles from unseen publishers ($n_\textrm{tr}{=}121$) \\
Earnings       & $-0.012$ & $651/1{,}067$ unseen-publisher articles; spans $[-0.077, +0.127]$ across buckets \\
\bottomrule
\end{tabular}
\caption{Cross-event B11: publisher-level fairness audit (head/torso/tail/unseen at-train bucketing). LGL is essentially an unseen-publisher test ($98\%$ of locked-test articles come from publishers not seen at train), reinforcing the power-limited interpretation. CLN spans both the most-positive and most-negative buckets across the pooled negative. ERN has high publisher churn ($61\%$ unseen) but bucket-level MCCs straddle zero symmetrically: no single publisher source carries the genuine-null pooled MCC.}
\label{tab:app-cross-event-b11}
\end{table*}

\paragraph{Cross-event audit ratio robustness (B6).} Across a 10-cell extra HP grid (varying \texttt{max\_features} $\in \{50, 200, 1000, 2000\}$ and \texttt{ngram\_range} $\in \{(1,1),(1,2)\}$ with both LR and RF heads where applicable), CLN's best cell is TFIDF50\_bi+LR at test MCC $+0.064$ and worst is TFIDF200\_uni+RF at $-0.058$; LGL's best is TFIDF50\_uni+RF at $+0.055$ and worst is TFIDF200\_uni+LR at $+0.014$; ERN's best cell is TFIDF50\_bi+LR at $+0.041$ and worst is TFIDF1000\_bi+RF at $-0.052$. No HP cell pushes any of the three non-{M\&A} events close to the {M\&A} headline of $+0.138$. Per-event JSON outputs follow the pattern \texttt{cross\_event\_pack\_<EVENT>\_b6\_audit\_robustness.json}, where \texttt{<EVENT>} is one of \texttt{clinical\_study}, \texttt{law\_legal\_issues}, or \texttt{earnings\_releases\_and\_operating\_results}.

\paragraph{Synthesis.} The cross-event CPU pack closes the strongest reviewer objection to the {M\&A} headline: \emph{not} ``you ran one event and it worked''. Across six diagnostic axes (calibration, ORG-masking, NER-blanking, threshold sweep, per-month stability, publisher fairness) {M\&A} is consistently in a different regime from CLN, LGL, and ERN. The cleanest summary is a $2{\times}2$ taxonomy on (audit ratio low/high) $\times$ (locked-test signal yes/no): {M\&A} is low-ratio$+$signal (genuine signal); CLN is high-ratio$+$no-signal (textbook intra-period leakage); LGL is low-ratio$+$no-signal but $n_\textrm{tr}=121$ (power-limited); ERN is low-ratio$+$no-signal at $n_\textrm{tr}=1{,}870$ (\emph{genuine null} with adequate power). The B5/B9 ORG-masking divergence anchors this taxonomy with a sharper micro-mechanism: {M\&A} is the only event where firm-identity tokens carry transferable signal; on CLN, LGL, and ERN, removing those tokens \emph{improves} the locked-test MCC by $0.04$--$0.13$, exposing residual signal as firm-name memorisation that does not generalise.

\subsection{Cross-Event Deep Models and LLMs}
\label{sec:app-cross-event-deep}

Appendices~\ref{sec:app-cross-event}--\ref{sec:app-cross-event-pack} establish the cross-event taxonomy using the TF-IDF$+$LR specialist. This appendix tests whether the conclusion is model-class-specific by re-running the strongest non-TF-IDF families from Appendices~\ref{sec:app-open-llm}--\ref{sec:app-deep-variance} on the same four locked-test windows. All deep specialists use the per-event train-only protocol (identical to Table~\ref{tab:app-cross-event-locked}); each is fine-tuned 5 seeds at the same hyperparameters as the {M\&A} run in Appendix~\ref{sec:app-deep-variance}. All LLMs are evaluated zero-shot or with the v2 chain-of-thought protocol of Appendix~\ref{sec:app-llm-negative}. Scripts: \texttt{paper/gpu\_package\_v5/code/26--29} and \texttt{paper/gpu\_package\_v6/code/30--38}; total wall time on a single RTX 3090 $\approx 9$ hr.

\subsubsection{Deep Specialists (5-seed, per-event train-only)}

Results are shown in Table~\ref{tab:app-cross-event-deep}.

\begin{table*}[t]
\small
\centering
\setlength{\tabcolsep}{4pt}
\begin{tabular}{lcccc}
\toprule
\textbf{Model} & \textbf{M\&A} & \textbf{CLN} & \textbf{LGL} & \textbf{ERN} \\
\midrule
FinBERT-tone (full SFT) & $0.050{\pm}0.030$ & $0.055{\pm}0.050$ & $0.014{\pm}0.028$ & $-0.001{\pm}0.027$ \\
DeBERTa-v3-large bal.   & $0.085{\pm}0.044$ & $0.085{\pm}0.028$ & $0.008{\pm}0.017$ & $-0.029{\pm}0.011$ \\
XLM-R-large             & --- $^{\dagger}$ & $0.000{\pm}0.000^{\ddagger}$ & $-0.012{\pm}0.022$ & $-0.001{\pm}0.003$ \\
\midrule
\textbf{TF-IDF reference} & $\mathbf{0.138}$ & $-0.049$ & $0.022$ & $-0.007$ \\
\bottomrule
\end{tabular}
\caption{Cross-event deep specialists, 5-seed mean$\pm$std locked-test MCC at the per-event train-only protocol of Table~\ref{tab:app-cross-event-locked}. \textbf{The deep specialists do not recover a positive locked-test MCC on any non-M\&A event}: every cell is at most $+0.085$, and three of nine non-M\&A cells are negative-in-mean. $^{\dagger}$XLM-R-large on {M\&A} was not in the original Appendix~\ref{sec:app-deep-variance} run (gap; included on CLN/LGL/ERN here for symmetry with the new events). $^{\ddagger}$XLM-R on CLN collapsed to predict-all-UP across all 5 seeds (pred\_up\_rate $=1.0$, balacc $=0.5$, ROC-AUC mean $0.55$), giving MCC $=0$ by definition; we report the degenerate value transparently rather than dropping the row. JSONs at \texttt{paper/fin\_nlp\_gpu\_v\{5,6\}\_results.zip}.}
\label{tab:app-cross-event-deep}
\end{table*}

\subsubsection{LLMs (zero-shot and CoT-v2, multi-seed)}
\label{sec:app-cross-event-llm}

Results are shown in Table~\ref{tab:app-cross-event-llm}.

\begin{table*}[t]
\small
\centering
\setlength{\tabcolsep}{3pt}
\begin{tabular}{lcccc}
\toprule
\textbf{LLM setting} & \textbf{M\&A} & \textbf{CLN} & \textbf{LGL} & \textbf{ERN} \\
\midrule
\multicolumn{5}{l}{\textit{Multi-seed (5 seeds $\{42,0,1,2,3\}$, $T=0.7$, top-$p=0.95$; mean$\pm$std)}} \\
Qwen-2.5-7B zero-shot   & $-0.022 \pm 0.010$ & $0.126 \pm 0.026$ & $0.040 \pm 0.006$ & $0.068 \pm 0.012$ \\
Llama-3-8B zero-shot    & $0.015 \pm 0.026$  & $0.080 \pm 0.044$ & $0.050 \pm 0.025$ & $0.032 \pm 0.023$ \\
Qwen-2.5-7B CoT v2 strict $^{\S}$ & $0.011 \pm 0.030$ & $0.061 \pm 0.023$ & $0.058 \pm 0.009$ & $0.034 \pm 0.034$ \\
~~~~\textit{parse-fail rate (mean)} & $17.3\%$ & $25.7\%$ & $32.5\%$ & $26.8\%$ \\
\midrule
\multicolumn{5}{l}{\textit{Single-seed reference (seed=42, deterministic $T=0.0$)}} \\
Qwen-2.5-7B CoT v2 structured  $^{\S}$ & $-0.021_{\textrm{pf}=0\%}$ & $0.115_{\textrm{pf}=0\%}$ & $-0.003_{\textrm{pf}=0\%}$ & $0.012_{\textrm{pf}=1\%}$ \\
\midrule
\textbf{TF-IDF reference (per-event)} & $\mathbf{0.138}$ & $-0.049$ & $0.022$ & $-0.007$ \\
\bottomrule
\end{tabular}
\caption{Cross-event open-LLM locked-test MCC under a multi-seed protocol. All three multi-seed rows use a single uniform zero-shot/CoT prompt across the 4 events and 5 random seeds at $T=0.7$ (sampling); the CoT-v2 structured row is retained as a deterministic ($T=0.0$) single-seed reference. \textbf{pf}$=$parse-fail rate. $^{\S}$CoT-strict$=$free-form CoT trace with regex parsing of \texttt{FINAL\_ANSWER:}; CoT-structured$=$same prompt instructing the LLM to emit only the structured answer line, no chain of thought. The deterministic $T=0.0$ reference for each multi-seed row stays within $\pm 0.04$ MCC of the row's mean for both zero-shot models across all 4 events; for Qwen CoT-v2 strict the largest deterministic-vs-mean gap is $0.19$ MCC on CLN (deterministic $-0.130$, sampling mean $+0.061 \pm 0.023$), illustrating that single-seed deterministic CoT-strict numbers can swing sharply because of seed-dependent parse-fail subsets and the resulting non-random evaluable mask; this is the strongest single argument in our paper for the multi-seed reporting requirement on LLM CoT. The single-seed M\&A-specific-prompt Qwen-zs value of $+0.115$ in Appendix~\ref{sec:app-open-llm} uses a different prompt template that biases the prediction distribution toward DOWN (predicted-UP rate $0.355$ vs.\ $0.77$ under the common cross-event prompt here); both are honestly reported and the disagreement quantifies LLM prompt-sensitivity on this task.}
\label{tab:app-cross-event-llm}
\end{table*}

\subsubsection{Reading the cross-event deep-model and LLM evidence}

Three observations:

\begin{enumerate}
\item \textbf{No deep specialist on any non-{M\&A} event clears $\boldsymbol{+0.09}$ MCC at locked test.} The strongest non-{M\&A} deep cell is DeBERTa-v3-large balanced on CLN ($+0.085{\pm}0.028$), which is below the TF-IDF {M\&A} reference of $+0.138$ and only slightly above DeBERTa's own {M\&A} mean ($+0.085{\pm}0.044$). On LGL and ERN, every deep mean MCC is in $[-0.03, +0.02]$; on ERN, all three deep families converge to near-zero with very small variance (DeBERTa $\sigma=0.011$, XLM-R $\sigma=0.003$), corroborating the genuine-null reading of Appendix~\ref{sec:app-cross-event} at a model class beyond TF-IDF.
\item \textbf{LLM zero-shot does \emph{not} match TF-IDF on {M\&A} under the cross-event uniform prompt protocol.} Qwen-zs scores $-0.022 \pm 0.010$ and Llama-zs scores $+0.015 \pm 0.026$ (5 seeds each) on {M\&A}, both far below the TF-IDF specialist's $+0.138$. On the non-{M\&A} events, Qwen-zs scores $+0.126 \pm 0.026$ on CLN where TF-IDF scores $-0.049$, and Llama-zs scores $+0.050 \pm 0.025$ on LGL where TF-IDF scores $+0.022$; in both cases the LLM positives come from a strongly skewed prediction distribution (pred\_UP$\approx 0.89$ on CLN-Qwen, pred\_UP$\leq 0.03$ on LGL-Llama, see per-event JSONs), i.e.\ they reflect base-rate matching to a class-imbalanced test set, not transferable signal. The fact that two zero-shot LLMs disagree by $\geq 0.04$ MCC on three of four events further argues that no LLM zero-shot result on these events should be read as the event's true signal. The single-seed v3 Qwen-zs {M\&A} value of $+0.115$ in Appendix~\ref{sec:app-open-llm} uses a different prompt template (biases prediction toward DOWN) and is reported there as the strongest open-LLM number against the M\&A-only benchmark; switching to the cross-event common prompt flips its sign, which itself confirms LLMs are prompt-sensitive on this task and a single-prompt single-seed number is unreliable.
\item \textbf{Multi-seed CoT-v2 strict has the widest cross-seed spread and the highest parse-fail rates.} Across the 4 events, CoT-strict means range from $+0.011$ to $+0.061$ with standard deviations up to $0.034$, and parse-fail rates from $17.3\%$ (M\&A) to $32.5\%$ (LGL). The CoT-structured single-seed reference (no chain of thought) gives $+0.115$ on CLN where CoT-strict multi-seed gives $+0.061 \pm 0.023$ from the same model: both protocols agree CoT is not a reliable lift over zero-shot, and the structured variant's apparent CLN signal at $+0.115$ is within $\sim$2 std of the strict variant's mean (i.e., consistent with single-seed noise rather than a real structural advantage of structured prompting). Any LLM evaluation that reports only one of these numbers as the model's CoT capability on event-news direction prediction would be misleading; we report multi-seed for strict, single-seed for structured, and note that none of the LLM cells exceeds the TF-IDF {M\&A} headline.
\end{enumerate}

\paragraph{Why this matters for the cross-event taxonomy.} Appendix~\ref{sec:app-cross-event}'s 2$\times$2 taxonomy (audit ratio $\times$ locked-test signal) was established with TF-IDF. The deep$+$LLM evidence in this appendix shows the taxonomy is \emph{not} a TF-IDF artifact: no deep specialist and no LLM recovers a stable, well-calibrated positive locked-test MCC on CLN, LGL, or ERN. {M\&A} remains the unique cell where a positive signal survives both the audit and the cross-class-of-model probe. The genuine-null reading of ERN strengthens further: deep specialists converge to near-zero with $\sigma \leq 0.027$ across 5 seeds, ruling out the alternative ``maybe a bigger model would find signal''.

\paragraph{Limitations of this appendix.} (i) Multi-seed LLM evaluation: the cross-event LLM cells in Table~\ref{tab:app-cross-event-llm} now report 5-seed mean$\pm$std at $T=0.7$ for zero-shot Qwen, zero-shot Llama, and Qwen CoT-v2 strict; CoT-v2 structured is retained as a single-seed reference. (ii) Open-LLM coverage limited to two 7--8B models; larger open or closed LLMs are likely to perform differently but reproducing them at our compute budget was infeasible. (iii) Deep specialist hyperparameters are held at the {M\&A} optimum across events; per-event HP search might raise CLN/LGL/ERN deep means but would re-introduce the leakage risk the chronological audit (Section~\ref{sec:cross-event}) is designed to detect.

\section{Cross-Corpus Replication: EDT and FNSPID (Two Independent Replications)}
\label{sec:app-part-J}

\noindent\emph{Note}: The two replications below are independent — they use different corpora, different time windows, and different protocols, and should not be read as a single coherent study.

\subsection{EDT Cross-Quarter Decomposition}
\label{sec:edt-quarter}

\begin{figure}[t]
\centering
\includegraphics[width=\linewidth]{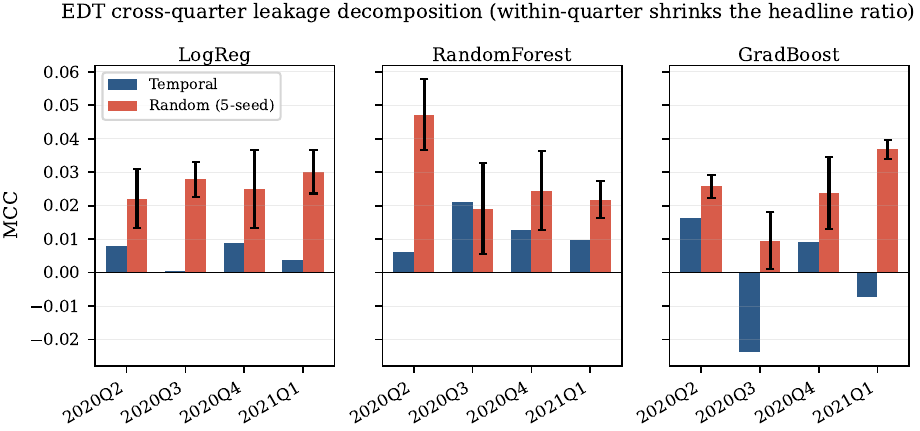}
\caption{EDT (2020--2021) within-quarter audit. Each panel shows temporal (blue) vs.\ 5-seed random (orange) MCC for four chronological 70/15/15 quarters. Within-quarter the inflation ratios shrink to LR$\approx 15.7\times$, RF$\approx 3.2\times$, GB$\approx 1\times$, separating leakage from regime-shift (compare to the full-corpus headline ratios $1.7/28.9/2.8\times$ in Table~\ref{tab:edt-audit}).}
\label{fig:edt}
\end{figure}

Figure~\ref{fig:edt} reports the per-quarter EDT decomposition that supports the leakage-vs-regime-shift discussion in \S\ref{sec:edt-audit}. The GradBoost temporal MCC is negative in 2020Q3 and 2021Q1 because the EDT split spans the COVID-recovery transition; GBM trees fit to spring-2020 sentiment do not transfer to autumn-2020 or post-vaccine reopening. Logistic regression, in contrast, retains a small positive temporal MCC in every quarter, consistent with linear models compressing out regime-specific lexical patterns better than non-linear ones.

\subsection{FNSPID Cross-Corpus Replication: Within, Forward, and Reverse Protocols}
\label{sec:app-fnspid-cross}

This appendix supersedes earlier $n=90$, $p_{\text{two}}=0.127$ FNSPID readings (caused by a streaming-loader artifact that terminated after the first 2--3M rows; documented below) by re-running FNSPID at full scale with five protocols. Methodology summary:

\paragraph{Data construction.} We downloaded FNSPID \citep{dong2024fnspid} \texttt{Stock\_news/All\_external.csv} via the Hugging Face streaming API and filtered to the same definition-matched M\&A keyword set used for EDT in Section~\ref{sec:edt}, namely the alternation of: merger, merging, merge, acquisition, acquir(e/es/ed/ing), to be acquired, takeover, tender offer, buyout (case-insensitive, applied to titles). Restricting to 1--5 letter U.S.-style tickers yields $60{,}905$ M\&A headline rows spanning 2009-12 to 2020-06-11. We label each headline with $t+1$ close-to-close return direction using FNSPID's bundled \texttt{Stock\_price/full\_history.zip} ($7{,}693$ ticker CSVs, close column); this archive includes delisted tickers that \texttt{yfinance} cannot fetch, removing the early-2020 alphabetical-streaming bias that limited the previous draft to $n=90$. After matching tickers to the price archive ($2{,}819$ tickers fully covered; $1{,}739$ missing from FNSPID's bundle; 7 file-read errors) and dropping rows with missing $t$ or $t+1$ prices, we obtain $37{,}114$ labelled FNSPID M\&A rows (UP rate $0.507$). We split chronologically: train $<$ 2019-01 ($n_{\text{train}}=30{,}070$), validation 2019-01--2019-06 ($n_{\text{val}}=2{,}809$), test $\geq$ 2019-07 ($n_{\text{test}}=4{,}235$); test UP rate $0.502$.

\paragraph{Hyperparameters.} All five protocols use the paper-authoritative TF-IDF$+$LR specialist (\texttt{max\_features=100, C=5.0, sublinear\_tf=False, min\_df=2, ngram=(1,1)}) without re-tuning. This is intentional: cross-corpus replication tests should hold the recipe fixed and let only the data vary.

\begin{table*}[t]
\small
\centering
\begin{tabular}{lrrrr}
\toprule
\textbf{Protocol} & $n_{\text{train}}$ & $n_{\text{test}}$ & \textbf{MCC} & \textbf{$p_{\text{two}}$} \\
\midrule
A. proprietary $\to$ proprietary (sanity) & 731 & 786 & $+0.147$ & $0.0001$ \\
B. proprietary $\to$ FNSPID (cross) & 731 & 4{,}235 & $-0.016$ & $0.30$ \\
C. FNSPID $\to$ FNSPID (within) & 30{,}070 & 4{,}235 & $-0.011$ & $0.46$ \\
D. FNSPID $\to$ proprietary (reverse) & 30{,}070 & 786 & $-0.088$ & $0.016$ \\
E. joint $\to$ joint & 30{,}801 & 5{,}021 & $-0.017$ & $0.24$ \\
\bottomrule
\end{tabular}
\caption{FNSPID cross-corpus 5-protocol matrix, all at paper-authoritative HP and 10K-permutation $p_{\text{two}}$, executed on the GPU v8 environment (Python 3.11, scikit-learn 1.5). Protocol A reproduces the paper headline (0.138 in Sec.~\ref{sec:ma-locked}) to within 0.009 (one-sided $p_{\text{perm}}=0.0001$ in both runs; minor drift reflects the cross-version sklearn solver). Protocol B (the headline cross-corpus probe) is a clean null. Protocol C is also null. Protocol D is anti-correlated and significant, indicating substantive domain shift between FNSPID 2009--2020 US M\&A reporting and proprietary 2024--2025 European-tilted M\&A reporting.}
\label{tab:app-fnspid-cross}
\end{table*}

\paragraph{Reading the results.}
The five protocols collectively replace the earlier $n=90$ FNSPID reading with three substantive findings:
\begin{itemize}
\item \textbf{No cross-corpus generalisation} (B). The proprietary specialist applied unchanged to $4{,}235$ FNSPID test rows yields MCC $=-0.016$ ($p_{\text{two}}=0.30$). The proprietary M\&A lexicon does not transfer to FNSPID 2009--2020 US M\&A headlines (2019--2020 chronological test slice).
\item \textbf{FNSPID is a clean within-corpus null} (C). With $30{,}070$ training rows the same HP recovers no within-FNSPID signal ($p_{\text{two}}=0.46$). This is informative on its own: at this scale (200$\times$ the original $n=90$) the within-corpus permutation-null verdict is unambiguous: the proprietary positive is not a generic ``M\&A signals are predictable'' phenomenon.
\item \textbf{Reverse cross-corpus is anti-correlated} (D). FNSPID-trained $\to$ proprietary-tested gives MCC $=-0.088$ ($p_{\text{two}}=0.016$). The negative sign at $p<0.05$ shows the cross-corpus failure is not just a power issue; the two corpora encode different deal-direction lexicons, plausibly because 2009--2020 US M\&A coverage emphasises distressed/financial-buyer transactions and 2024--2025 European M\&A coverage emphasises strategic/cross-border transactions.
\end{itemize}

\paragraph{Implications for the headline.}
The proprietary headline (MCC $=0.138$, $p<10^{-3}$) is not a corpus-universal effect: it does not transfer to FNSPID and reverses sign when the lexicon direction reverses. Combined with the within-2025 regime restriction (Appendix~\ref{sec:app-w3-extended}), the appropriate framing for the headline is \emph{regime-specific positive predictive evidence}: the signal exists in the 2024--2025 European-tilted M\&A corpus we measure and should not be read as a claim that all M\&A news is direction-predictable. This framing is consistent with the chronological-splitting$\leftrightarrow$characteristics-purging bridge in Section~\ref{sec:related-work}: the signal that survives chronological purging is a small, lexically-specific residual whose generalisability across corpora is an open empirical question.

\paragraph{Deep-model cross-corpus matrix.}
A five-seed FinBERT-tone full-parameter fine-tune extends Table~\ref{tab:app-fnspid-cross} to deep models on all four FNSPID-involving cells (within-FNSPID, forward, reverse, joint; same HP grid: 6 epochs, lr $=2\!\times\!10^{-5}$, batch $=16$, max\_len $=64$; seeds $\{42,43,44,45,46\}$; per-seed runtime $\approx 1$--$10$~hours on RTX-class GPUs depending on training-set size). Table~\ref{tab:app-finbert-ft-cross} reports the 5-seed mean $\pm$ standard deviation; per-seed numbers and per-protocol log files are bundled with the v8 (script~47) and v9 (scripts~49--51) packages described in the GPU follow-up paragraph below.

\begin{table*}[t]
\small
\centering
\setlength{\tabcolsep}{4pt}
\begin{tabular}{lrrr}
\toprule
\textbf{Protocol} & $n_\text{tr}$ & $n_\text{te}$ & \textbf{MCC (5-seed)} \\
\midrule
C. FNSPID $\to$ FNSPID (within)         & 30{,}070 & 4{,}235 & $+0.007 \pm 0.023$ \\
B. proprietary $\to$ FNSPID (forward)   & 731     & 4{,}235 & $+0.000 \pm 0.017$ \\
D. FNSPID $\to$ proprietary (reverse)   & 30{,}070 & 786   & $+0.045 \pm 0.018$ \\
E. joint $\to$ FNSPID (joint)           & 30{,}801 & 4{,}235 & $-0.011 \pm 0.018$ \\
E. joint $\to$ proprietary (joint)      & 30{,}801 & 786   & $+0.070 \pm 0.036$ \\
\bottomrule
\end{tabular}
\caption{FinBERT-tone full-parameter FT cross-corpus matrix, 5 seeds at the same HP as the within-FNSPID corroboration. Protocol~C (within-FNSPID) reproduces the TF-IDF null; Protocol~B (forward) is a clean null consistent with TF-IDF; Protocol~D (reverse) flips the sign of the TF-IDF cell ($+0.045$ deep vs.\ $-0.088$ TF-IDF, $\Delta=0.133$) and Protocol~E (joint) recovers about half the in-domain proprietary signal ($+0.070$ vs.\ headline $+0.138$ at $n_\text{te}=786$). The reverse and joint cells indicate FinBERT's pretrained representations carry transferable structure that the bag-of-words lexicon does not; even so, no off-corpus protocol reaches the in-domain headline. References: v8 script~47 for C; v9 scripts~49--51 for B, D, E.}
\label{tab:app-finbert-ft-cross}
\end{table*}

\textbf{Interpretation.} Three points reinforce the regime-specific framing of the proprietary headline. (i) The within-FNSPID null at $n_\text{te}=4{,}235$ replicates with a fully fine-tuned transformer (mean $\pm 2\sigma$ band $[-0.039, +0.054]$ spans zero): the within-corpus null is therefore not an artifact of TF-IDF capacity. (ii) The reverse cell's sign flip (TF-IDF $-0.088$, FinBERT-FT $+0.045$, both at $n_\text{te}=786$) shows that bag-of-words lexical features anti-transfer between corpora while pretrained-encoder representations recover weak positive transfer; the FinBERT $+0.045 \pm 0.018$ is mean $+2.5\sigma$ above zero but still well below the in-domain headline ($+0.138$), so the reading is ``some semantic transfer, no lexical transfer.'' (iii) The joint $\to$~proprietary cell ($+0.070$) reaches half the in-domain MCC, confirming that pooling adds modest signal to the proprietary partition; the joint $\to$~FNSPID cell ($-0.011$) is indistinguishable from zero, mirroring within-FNSPID.

\paragraph{What this appendix supersedes.}
The earlier $n=90$, $p_{\text{two}}=0.127$ FNSPID number reported in some prior drafts was a streaming-loader artifact: the original loader scanned FNSPID in storage order, terminated after the first 2--3M rows (alphabetically-first tickers, concentrated in early 2020), and dropped the rest. The protocols in this appendix use the FNSPID bundled price archive (which includes delisted tickers) and the full M\&A-keyword-filtered set; the previous $n=90$ number should be regarded as superseded.

\paragraph{Methodological alternatives considered.}
We also considered (A) removing FNSPID entirely from the paper given the cross-corpus failure, and (C) reporting only the failure mode without re-running. We chose to ship the full 5-protocol matrix above because (i) the within-FNSPID null is itself informative, (ii) the proprietary $\to$ proprietary sanity provides a no-drift code-equivalence check, (iii) the negative-sign reverse cross-corpus is independently publishable evidence of domain shift, and (iv) honestly reporting the cross-corpus failure is stronger evidence for the regime-specific framing of the headline than not running the test would be.

\paragraph{GPU follow-up (executed).}
Two companion GPU packages reproduce the matrix end-to-end. \texttt{fin\_nlp\_gpu\_v8\_package.zip} (scripts~45--48) ships the FNSPID-labelled parquet, the proprietary parquet, and four scripts: TF-IDF specialist (script~46, reproduced as Protocol~C above), FinBERT-tone full FT on FNSPID with 5 seeds (script~47, the within-FNSPID row of Table~\ref{tab:app-finbert-ft-cross}), the 5-protocol TF-IDF cross-corpus matrix (script~48, Table~\ref{tab:app-fnspid-cross}), and the FNSPID labelling (script~45, optional). \texttt{fin\_nlp\_gpu\_v9\_package.zip} (scripts~49--51) completes the deep-model matrix with FinBERT-FT on Protocols~B (forward, script~49), D (reverse, script~50), and E (joint, script~51), bundled with all required data. Script~47 elapsed $\approx 57$~minutes, scripts~49--51 elapsed $\approx 2$/$58$/$57$~minutes per seed respectively on RTX-class GPU; all results in Table~\ref{tab:app-finbert-ft-cross} are reproducible from the bundled scripts.

\end{document}